\documentclass{article}

\usepackage{microtype}
\usepackage{graphicx}
\usepackage{booktabs} 

\usepackage{hyperref}

\usepackage[accepted]{icml2024}

\usepackage{amsmath}
\usepackage{amssymb}
\usepackage{mathtools}
\usepackage{amsthm}

\usepackage[capitalize,noabbrev]{cleveref}

\theoremstyle{plain}
\newtheorem{theorem}{Theorem}[section]

\theoremstyle{definition}
\newtheorem{definition}[theorem]{Definition}

\theoremstyle{remark}

\usepackage[textsize=tiny]{todonotes}

\usepackage{amsmath,amsfonts,bm}

\def\eqref#1{equation~\ref{#1}}
\def\Eqref#1{Equation~\ref{#1}}

\def\1{\bm{1}}

\def\rva{{\mathbf{a}}}

\def\rvu{{\mathbf{i}}}

\def\rvk{{\mathbf{k}}}

\def\rvp{{\mathbf{p}}}
\def\rvq{{\mathbf{q}}}

\def\rvs{{\mathbf{s}}}

\def\rvu{{\mathbf{u}}}
\def\rvv{{\mathbf{v}}}

\def\rvx{{\mathbf{x}}}

\def\mA{{\bm{A}}}

\def\mC{{\bm{C}}}

\def\mK{{\bm{K}}}

\def\mT{{\bm{T}}}

\DeclareMathAlphabet{\mathsfit}{\encodingdefault}{\sfdefault}{m}{sl}
\SetMathAlphabet{\mathsfit}{bold}{\encodingdefault}{\sfdefault}{bx}{n}

\def\gB{{\mathcal{B}}}

\def\gD{{\mathcal{D}}}

\def\gL{{\mathcal{L}}}

\def\gN{{\mathcal{N}}}

\def\gS{{\mathcal{S}}}
\def\gT{{\mathcal{T}}}

\def\gX{{\mathcal{X}}}
\def\gY{{\mathcal{Y}}}

\newcommand{\E}{\mathbb{E}}

\newcommand{\R}{\mathbb{R}}

\newcommand{\src}{s}
\newcommand{\tgt}{t}
\newcommand{\srcd}{\vecmu}
\newcommand{\tgtd}{\vecnu}
\newcommand{\srcT}{T_{\src}}
\newcommand{\tgtT}{T_{\tgt}}
\newcommand{\srct}{\mathcal{T}_\text{src}}
\newcommand{\tgtt}{\mathcal{T}_\text{tgt}}

\newcommand{\Wd}{Wasserstein distance\xspace}
\newcommand{\GW}{Gromov-Wasserstein\xspace}
\newcommand{\GWD}{Gromov-Wasserstein Distance\xspace}
\newcommand{\GWd}{Gromov-Wasserstein distance\xspace}

\newcommand{\matA}{\mA}
\newcommand{\matC}{\mC}
\newcommand{\matK}{\mK}
\newcommand{\matT}{\mT}
\newcommand{\vecu}{\rvu}
\newcommand{\vecv}{\rvv}
\newcommand{\vecs}{\rvs}
\newcommand{\veca}{\rva}

\newcommand{\vecx}{\rvx}

\newcommand{\vecp}{\rvp}
\newcommand{\vecq}{\rvq}
\newcommand{\veck}{\rvk}

\newcommand{\ourmethod}{PEARL\xspace}
\newcommand{\CPA}{CPA\xspace}

\newcommand{\PT}{Preference Transformer\xspace}

\newcommand{\OT}{Optimal Transport\xspace}

\newcommand{\ot}{optimal transport\xspace}
\newcommand{\pbRL}{preference-based RL\xspace}

\newcommand{\oral}{F_{\text{oracle}}\xspace}
\newcommand{\cpal}{F_{\text{CPA}}\xspace}

\def\vecmu{{\bm{\mu}}}
\def\vecnu{{\bm{\nu}}}

\newcommand{\mean}[1]{#1}
\newcommand{\bmean}[1]{\textbf{#1}}
\newcommand{\std}[1]{\scalebox{.70}{$\pm$ #1}}

\usepackage{enumitem} 
\usepackage{subfig} 
\usepackage{multirow}
\usepackage{xspace}
\usepackage{makecell} 
\usepackage{xcolor}
\usepackage{colortbl}

\usepackage{algorithm}
\renewcommand{\algorithmicrequire}{\textbf{Input:}} 
\renewcommand{\algorithmicensure}{\textbf{Output:}}

\graphicspath{{figs/}}

\icmltitlerunning{Cross-task Preference Alignment and Robust Reward Learning for Robotic Manipulation}

\begin{document}

\twocolumn[
\icmltitle{PEARL: Zero-shot Cross-task Preference Alignment and Robust Reward Learning for Robotic Manipulation}




\begin{icmlauthorlist}
\icmlauthor{Runze Liu}{thu}
\icmlauthor{Yali Du}{kcl}
\icmlauthor{Fengshuo Bai}{bigai,sjtu}
\icmlauthor{Jiafei Lyu}{thu}
\icmlauthor{Xiu Li}{thu}
\end{icmlauthorlist}

\icmlaffiliation{thu}{Tsinghua Shenzhen International Graduate School, Tsinghua University}
\icmlaffiliation{kcl}{King’s College London}
\icmlaffiliation{bigai}{Beijing Institute for General AI}
\icmlaffiliation{sjtu}{Shanghai Jiao Tong University}

\icmlcorrespondingauthor{Yali Du}{yali.du@kcl.ac.uk}
\icmlcorrespondingauthor{Xiu Li}{li.xiu@sz.tsinghua.edu.cn}

\icmlkeywords{Machine Learning, ICML}

\vskip 0.3in
]



\printAffiliationsAndNotice{}  

\begin{abstract}
In preference-based Reinforcement Learning (RL), obtaining a large number of preference labels are both time-consuming and costly. Furthermore, the queried human preferences cannot be utilized for the new tasks. In this paper, we propose Zero-shot Cross-task Preference Alignment and Robust Reward Learning (PEARL), which learns policies from cross-task preference transfer without \textbf{any} human labels of the target task. Our contributions include two novel components that facilitate the transfer and learning process. The first is Cross-task Preference Alignment (CPA), which transfers the preferences between tasks via optimal transport. The key idea of CPA is to use Gromov-Wasserstein distance to align the trajectories between tasks, and the solved optimal transport matrix serves as the correspondence between trajectories. The target task preferences are computed as the weighted sum of source task preference labels with the correspondence as weights. Moreover, to ensure robust learning from these transferred labels, we introduce Robust Reward Learning (RRL), which considers both reward mean and uncertainty by modeling rewards as Gaussian distributions. Empirical results on robotic manipulation tasks from Meta-World and Robomimic demonstrate that our method is capable of transferring preference labels across tasks accurately and then learns well-behaved policies. Notably, our approach significantly exceeds existing methods when there are few human preferences. The code and videos of our method are available at:~\url{https://sites.google.com/view/pearl-preference}.
\end{abstract}

\section{Introduction}

\begin{figure*}[!t]
\centering
\includegraphics[width=0.9\linewidth]{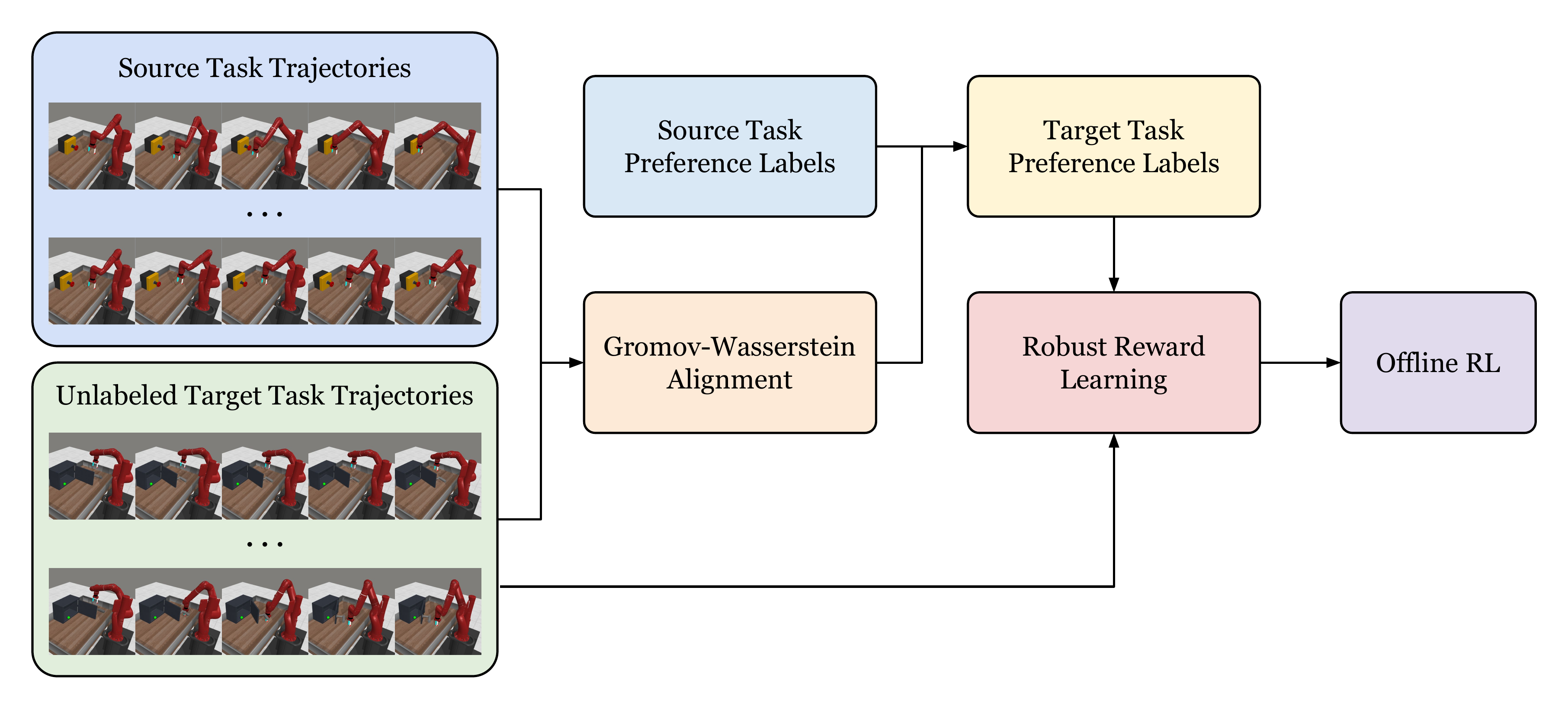}
\caption{Framework of \ourmethod. Given unlabeled target task trajectories and source task trajectories and their preference labels, the trajectories between tasks are first aligned via Gromov-Wasserstein distance. Then the target task preference labels are computed by the solved optimal transport matrix and source task preference labels. The reward model is learned robustly and finally offline RL algorithm is applied to obtain the policy.
}
\label{fig:framework}
\end{figure*}

In recent years, great achievements have been made in Reinforcement Learning, particularly in solving sequential decision-making problems given a well-defined reward function~\citep{mnih2013playing, silver2016mastering, VinyalsBCMDCCPE19, abs-1912-06680, PiCor}. However, the practical deployment of RL algorithms is often hindered by the extensive effort and time required for reward engineering. Additionally, there are risks of reward hacking, where RL agents exploit reward functions in unanticipated ways, leading to unexpected and potentially unsafe outcomes. Moreover, the challenge of aligning RL agents with human values, particularly sensorimotor skills as discussed in this paper, through crafted reward functions remains substantial in real-world applications.

As a promising alternative, preference-based RL~\citep{christiano2017deep} introduces a paradigm different from traditional RL by learning reward functions via human preferences between trajectories rather than manually designed reward functions. By directly capturing human intentions, preference-based RL has demonstrated an ability to teach agents novel behaviors that align more closely with human values. However, while the strides made in \pbRL are significant~\citep{park2022surf, liang2022reward, liu2022meta}, current algorithms come with their own set of challenges. First, they are heavily reliant on a vast number of online queries to human experts for preference labels for reward and policy learning. This dependency not only increases the time and cost associated with training but also results in data that cannot be recycled or repurposed for new tasks. Each new task encountered demands its own set of human preference labels, creating a cycle of labeling that is both resource-intensive and inefficient. While~\citet{hejna2023few} leverage prior data to pre-train reward functions via meta-learning and adapts quickly with new task preference data, the need for millions of pre-collected preference labels and further online queries makes this approach impractical in many scenarios.

Recently, \GW~(GW) distance~\citep{memoli2011gromov} has shown effectiveness in a variety of structured data matching problems, such as graphs~\citep{xu2019gromov} and point clouds~\citep{peyre2016gromov}. \GWd measures the relational distance and finds the optimal transport plan across different domains. Inspired by this, we consider using \GWd as an alignment tool between the trajectories of source and target tasks. Given two sets of trajectories from source and target tasks respectively, we can identify the corresponding trajectory pairs between tasks based on the solved optimal transport matrix. Hence, a zero-shot cross-task \pbRL algorithm can be developed that utilizes previously annotated preference data to transfer the preferences across tasks.

In this work, we aim to leverage data collected from existing source tasks to reduce the human labeling cost for unseen target tasks. We propose to use \GWd to find the correspondence between trajectories from source tasks and target tasks and compute preference labels according to trajectory alignment. Our method only requires a small number of preference labels from source tasks, then obtaining abundant preference labels for the target task. However, the transferred labels may contain a proportion of incorrect labels, which significantly affect reward and policy learning. To learn robustly from \CPA labels, we introduce a novel distributional reward modeling approach, which not only captures the average reward but also measures the reward uncertainty. The framework of our method is shown in Figure~\ref{fig:framework}.

In summary, our contributions are three-fold. First, we introduce \textbf{C}ross-task \textbf{P}reference \textbf{A}lignment~(CPA), the \textbf{first} zero-shot cross-task \pbRL approach that utilizes small amount of preference data from previous tasks to infer pseudo labels via optimal transport. Second, we propose \textbf{R}obust \textbf{R}eward \textbf{L}earning~(RRL), to ensure robust learning from \CPA labels. Last, we validate the effectiveness of our approach through experiments on several robotic manipulation tasks of Meta-World~\citep{yu2020meta} and Robomimic~\citep{mandlekar2021matters}. The empirical results show the strong abilities of our method in zero-shot preference transfer. Moreover, it is shown that our method significantly outperforms current methods when there is a lack of human preference labels.

\section{Related Work}

\paragraph{Preference-based Reinforcement Learning.}

Preference-based RL algorithms have achieved great success by aligning with human feedback~\citep{christiano2017deep, ibarz2018reward, lee2021bpref, ouyang2022training, bai2022training}. The main challenge of \pbRL is feedback efficiency and many recent \pbRL works have contributed to tackle this problem. To improve feedback efficiency, PEBBLE~\citep{lee2021pebble} proposes to use unsupervised exploration for policy pre-training. SURF~\citep{park2022surf} infers pseudo labels based on reward confidence to take advantage of unlabeled data, while RUNE~\citep{liang2022reward} facilitates exploration guided by reward uncertainty. Meta-Reward-Net~\citep{liu2022meta} further improves the efficiency by incorporating the performance of the Q-function during reward learning. SEER~\citep{SEER} proposes to utilize label smoothing and policy regularization via conservatism to tackle this issue. However, most current \pbRL methods still requires a large number of human preference labels for training new tasks, and the data cannot be utilized for learning other tasks. To leverage preference data on source tasks and reducing the amount of human feedback,~\citet{hejna2023few} leverage meta learning to pre-train the reward function, achieving fast adaptation on new tasks with few human preferences. Despite the success of~\citet{hejna2023few} in reducing human cost, it still needs 1.5 million labels for pre-training and further online querying for the new task. Recently, offline \pbRL draws more attention with the increasing interests in offline RL~\citep{kostrikov2022offline, SEABO}. \PT~(PT)~\citep{kim2023preference} proposes to use Transformer architecture to model non-Markovian rewards and outperforms previous methods that model Markovian rewards. IPL~\citep{hejna2023inverse} learns policies without reward functions. Nonetheless, PT and IPL still require hundreds of human labels. Our method differs from prior methods that we only need a small number of human labels from source tasks and can obtain extensive preference labels for the new task.

\paragraph{Optimal Transport.}

\OT~(OT) has been widely studied in domain adaptation~\citep{damodaran2018deepjdot, shen2018wasserstein}, graph matching~\citep{titouan2019optimal, xu2019gromov}, recommender systems~\citep{li2022gromov}, and imitation learning~\citep{fickinger2022cross}. For example, GWL~\citep{xu2019gromov} is proposed to jointly learn node embeddings and perform graph matching.~\citet{li2022gromov} transfer the knowledge from the source domain to the target domain by using \GWd to align the representation distributions. In the context of RL, there are several imitation learning methods that utilize OT to align the agent's and expert's state-action distributions~\citep{dadashi2021primal, cohen2021imitation, haldar2023watch, luo2023optimal, haldar2023teach}. For cross-task imitation learning method, GWIL~\citep{fickinger2022cross} aligns agent states between source and target tasks and computes pseudo rewards based on the solved \ot plan. \ourmethod is the first \pbRL algorithm that leverages \ot for cross-task alignment. \ourmethod does not perform representation space alignment, which requires additional gradient computation. It directly uses \GWd to align trajectory distributions between tasks and compute preference labels for the target tasks according to the transport matrix.

\paragraph{Distributional Modeling for Robust Learning from Noisy Samples.}

Traditional representation learning techniques extract features as fixed points. However, such modeling fails to adequately capture data uncertainty, leading to suboptimal performance with noisy data. A series of studies have proposed modeling features as distributions to enhance robustness, seen in person Re-ID~\citep{yu2019robust}, face recognition~\citep{chang2020data}, scene graph generation~\citep{yang2021probabilistic}, Vision-Language Pre-training (VLP)~\citep{ji2022map}. Specifically, these methods utilize Gaussian distributions rather than fixed points to model features, interpreting variance as uncertainty. In \pbRL,~\citet{xue2023reinforcement} proposes an encoder-decoder architecture for reward modeling, which encodes state-action features as Gaussian distributions. Consequently, the features can be manipulated in a latent space and they are constrained to be close to a prior distribution to stabilize reward learning process. In our work, we model reward distributions rather than feature distributions and we are the first to model reward distribution in \pbRL to the best of our knowledge.

\begin{figure*}[!t]
\centering
\includegraphics[width=0.85\linewidth]{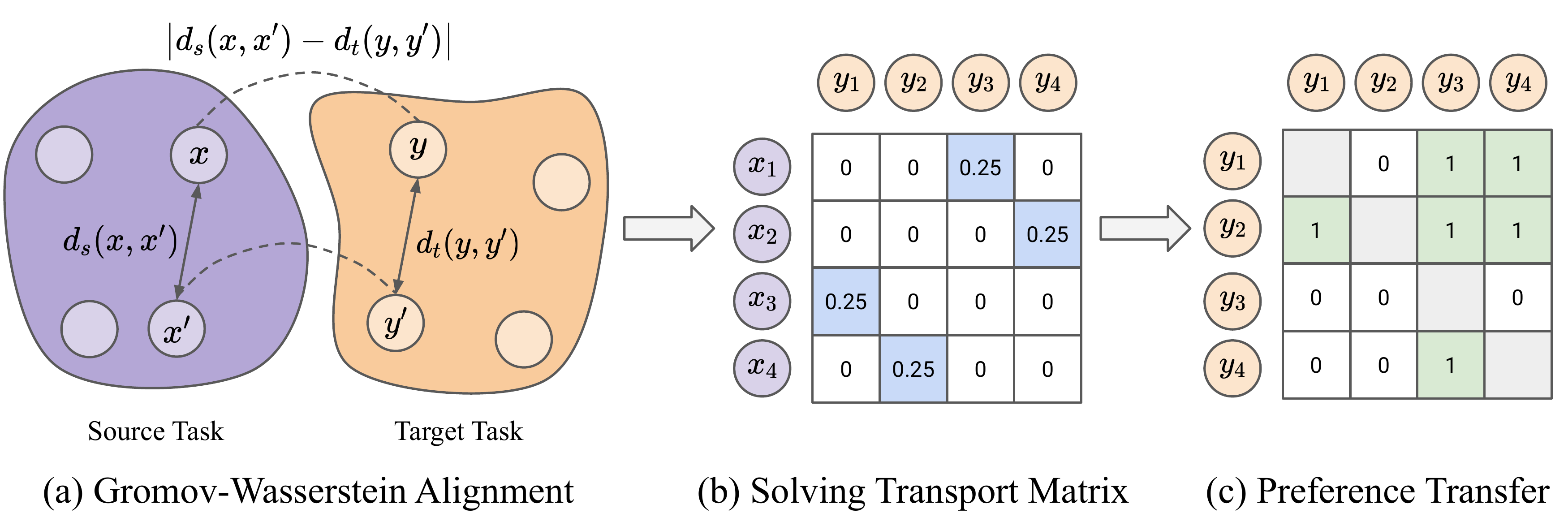}
\caption{Diagram of cross-task preference alignment. The circle $\bigcirc$ represents a trajectory segment in each task. (a) \CPA uses \GWd as a relational distance metric to align trajectory distributions between source and target tasks. (b) The optimal transport matrix is solved by optimal transport, with each element representing the correspondence between trajectories of two tasks. (c) The preference labels of trajectory pairs of the target task are computed based on trajectory correspondence by~\Eqref{eq:CPA}. $z(y_1,y_3)=1$ indicates that $y_3$ is better than $y_1$ and $0$ indicates $y_1$ is preferred.}
\label{fig:CPA}
\end{figure*}

\section{Problem Setting \& Preliminaries}

\paragraph{Problem Setting.}
In this paper, we consider preference transfer between tasks share the same action space. We assume there exists a task distribution $p(\mathcal{T})$, with each task $\mathcal{T}$ corresponding to a distinct Markov Decision Process (MDP). MDP is defined by the tuple $(\mathcal{S}, \mathcal{A}, \mathcal{P}, \mathcal{R}, \gamma)$ consisting of a state space $\mathcal{S}$, an action space $\mathcal{A}$, a transition function $\mathcal{P}: \mathcal{S} \times \mathcal{A} \to \mathcal{S}$, a reward function $\mathcal{R}: \mathcal{S} \times \mathcal{A} \to \mathbb{R}$, and a discount factor $\gamma \in [0,1)$. While the action space $\mathcal{A}$ remain identical across these MDPs, the state space $\mathcal{S}$, the transition function $\mathcal{P}$, the reward function $\mathcal{R}$, and the discount factor $\gamma$ can differ.

In this context, our paper introduces the problem of zero-shot preference transfer. We consider a source task $\srct \sim p(\mathcal{T})$ and a target task $\tgtt \sim p(\mathcal{T})$. Assume we have $M$ trajectories {$x_i$} of task $\srct$, $i=1, \cdots, M$, along with preference labels of all combinations of trajectory pairs $(x_i, x_{i^\prime})$ where $i, i^\prime=1, \cdots, M, i < i^\prime$. For task $\tgtt$, there are $N$ trajectories {$y_j$}, $j=1, \cdots, N$. The goal of our method is to learn a policy $\pi(a \mid s)$ for task $\tgtt$ with preference labels transferred from task $\srct$.

\paragraph{Preference-based Reinforcement Learning.}
Preference-based RL is assumed to have no access to the ground-truth reward function and learns a reward function $\widehat{r}_\psi$ from human preferences. A trajectory segment of length $H$ is represented as $x=\{\vecs_1, \veca_1, \cdots, \vecs_H, \veca_H\}$. Given a pair of segments $(x^0, x^1)$, a human provides a preference label $z \in \{0, 1, 0.5\}$, where $0$ indicates that $x^0$ is preferred over $x^1$ (denoted as $x^0 \succ x^1$), $1$ denotes the reverse preference, and $0.5$ indicates the two segments are equally preferable. The preference predictor formulated via the Bradley-Terry model~\citep{bradley1952rank} is:
\begin{equation}
    P_\psi[x^0 \succ x^1] = \frac{\exp\sum_t \widehat{r}_\psi(\vecs_t^0, \veca_t^0)}{\exp\sum_t \widehat{r}_\psi(\vecs_t^0, \veca_t^0) + \exp\sum_t \widehat{r}_\psi(\vecs_t^1, \veca_t^1)}.
\end{equation}
With a dataset containing trajectory pairs and their labels $\gD = \{(x^0, x^1, z)\}$, the parameters of the reward function can be optimized using the following cross-entropy loss:
\begin{equation}
\begin{aligned}
    \gL_{\text{ce}}(\psi) = - \underset{{(x^0,x^1,z)\sim\mathcal{D}}}{\E} \Big[ &(1-z)\log P_\psi[x^0 \succ x^1] \\
    &+ z\log P_\psi[x^1 \succ x^0] \Big].
\end{aligned}
\label{eq:sup_loss}
\end{equation}
By aligning the reward function with human preferences, the policy can be learned from labeled transitions by $\widehat{r}_\psi$ via RL algorithms.

\paragraph{Optimal Transport.}

\OT~(OT) aims to find the optimal coupling of transporting one distribution into another with minimum cost. Unlike \Wd, which measures absolute distance, \GWd is a relational distance metric incorporating the metric structures of the underlying spaces~\citep{memoli2011gromov, peyre2016gromov}. Besides, \GWd measures the distance across different domains, which is beneficial for cross-domain learning. The mathematical definition of \GWd is as follows:

\begin{definition}
(\GWD~\citep{peyre2016gromov}) Let $(\gX, d_\gX, \mu_\gX)$ and $(\gY, d_\gY, \mu_\gY)$ denote two metric measure spaces, where $d_\gX$ and $d_\gY$ represent distance metrics measuring similarity within each task, and $\mu_\gX$ and $\mu_\gY$ are Borel probability measures on $\gX$ and $\gY$, respectively. For $p \in [1, \infty)$, the \GWd is defined as:
\begin{equation}
    \inf_{\gamma \in \Pi(\mu_\gX, \mu_\gY)} \underset{\gX\times\gY, \gX\times\gY}{\iint} L(x, x^\prime, y, y^\prime)^p d \gamma(x, y) d \gamma(x^\prime, y^\prime),
\end{equation}
where $L(x, x^\prime, y, y^\prime) = \left| d_\gX(x, x^\prime) - d_\gY(y, y^\prime) \right|$ denotes the relational distance function, and $\Pi(\mu_\gX, \mu_\gY)$ is the set of joint probability distributions with marginal distributions $\mu_\gX$ and $\mu_\gY$. $p$ is assumed to be $2$ in the paper.
\end{definition}

\begin{figure*}[!t]
\centering
\includegraphics[width=0.7\linewidth]{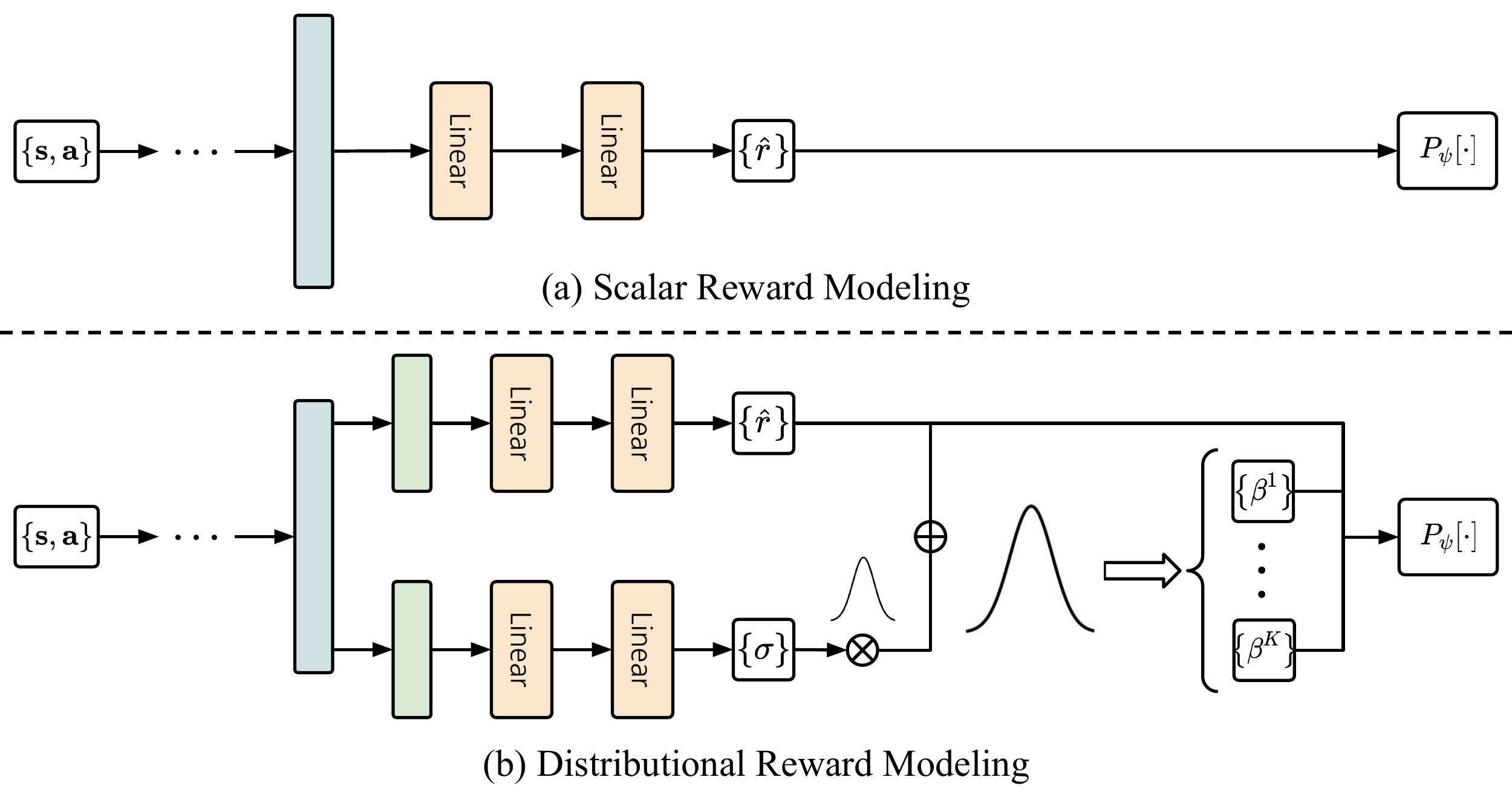}
\caption{Different types of reward modeling. (a) Scalar reward modeling, which only considers scalar rewards. This modeling type is widely used in \pbRL algorithms~\citep{christiano2017deep, lee2021pebble, kim2023preference}. (b) Distributional reward modeling, which adds a branch for modeling reward uncertainty in addition to reward mean.}
\label{fig:RPT}
\end{figure*}

\section{Method}

In this section, we present \ourmethod, a zero-shot cross-task \pbRL algorithm that transfers preferences between tasks and learns robustly without any human labels. First, we propose Cross-task Preference Alignment to align the trajectories of source and target tasks using \ot and computes preference labels according to the solved optimal alignment matrix. Second, we introduce Robust Reward Learning, which additionally incorporates the reward uncertainty by modeling the rewards from a distributional perspective, enabling robust learning from noisy labels.

\subsection{Cross-task Preference Alignment}

\GWd shows great abilities in aligning structural information, such as correspondence of edges between two graphs. Therefore, we consider using \GWd as an alignment metric between the trajectories of source and target tasks, finding the alignment of paired trajectories between tasks, and inferring preference labels based on the correspondence and preference labels of the source trajectory pairs. The diagram of CPA is shown in Figure~\ref{fig:CPA}.

\CPA aims to identify the correspondence between two sets of trajectories and transfer the preferences accordingly. In this paper, we consider preference transfer problem from a source task $\gS$ to a target task $\gT$, with their distributions denoted as $\srcd$ and $\tgtd$, respectively. Assume we have $M$ segments with pairwise preference labels $\{x_{i}\}_{i=1}^M$ from the source task and $N$ segments $\{y_{j}\}_{j=1}^{N}$ from the target task. The trajectories can be represented by the probability measures $\srcd = \sum_{i=1}^M u_i \delta_{x_i}$ and $\tgtd = \sum_{j=1}^N v_j \delta_{y_j}$, where $\delta_{x}$ denotes the Dirac function centered on $x$. The weight vectors $\{u_i\}_{i=1}^M$ and $\{v_j\}_{j=1}^N$ satisfy $\sum_{i=1}^M u_i=1$ and $\sum_{j=1}^N v_j=1$, respectively. The empirical \GWd for aligning trajectories between source and target tasks is expressed as:
\begin{equation}
    \min_{\matT \in \Pi(\srcd, \tgtd)}
    \sum_{i,i^{\prime},j,j^{\prime}}
    \left|d_\src(x_i, x_{i^\prime}) - d_\tgt(y_j, y_{j^\prime})\right|^2 T_{ij} T_{i^{\prime} j^{\prime}},
\label{eq:emperical_GW}
\end{equation}
where the \ot matrix is $\matT=[T_{ij}]$, $\Pi(\srcd, \tgtd)$ denotes the set of all couplings between $\srcd$ and $\tgtd$, $\Pi(\srcd, \tgtd) = \{ \matT \in \mathbb{R}^{M \times N} \mid \matT \mathbf{1}_{N}=\srcd, \matT^\top \mathbf{1}_{M}=\tgtd \}$, $\mathbf{1}_{M}$ denotes a $M$-dimensional vector with all elements equal to one, and $d_s, d_t$ represent the distance function in each task, such as Euclidean function or Cosine function.

Upon solving~\Eqref{eq:emperical_GW}, we obtain the \ot matrix $\matT$ representing the correspondence between the trajectories of the two tasks. Each element, $T_{ij}$, indicates the probability that trajectory $x_i$ matches trajectory $y_j$, and the $j$-th column represents the correspondence between $y_j$ and all source trajectories. Therefore, for a pair of trajectories $(y_j, y_j^\prime)$, we can identify the paired relations based on the \ot matrix. We define the trajectory pair matching matrix $\matA^{j j^\prime}$ for each $(y_j, y_j^\prime)$ by multiplying the $j$-th column $\matT_{\cdot j}$ and the transpose of $j^\prime$-th column $\matT_{\cdot j^\prime}^\top$:
\begin{equation}
    \matA^{j j^\prime} = \matT_{\cdot j} \matT_{\cdot j^\prime}^\top,
\label{eq:matA}
\end{equation}
where $\matA^{j j^\prime} \in \R^{N \times N}$, and each element $A_{i i^\prime}^{j j^\prime}$ of the matrix represents the correspondence of trajectory pair $(y_j, y_j^\prime)$ with trajectory pair $(x_i, x_i^\prime)$ from the source task. If we denote the preference label of $(x_i, x_i^\prime)$ as $z(x_i, x_{i^\prime})$, then the \ourmethod label of $(y_i, y_{i^\prime})$ is computed as follows:
\begin{equation}
    z(y_j, y_{j^\prime}) = \sum_{i} \sum_{i^\prime \neq i} A_{ii^\prime}^{j j^\prime} z(x_i, x_{i^\prime}),
\label{eq:CPA}
\end{equation}
where $i^\prime \neq i$ is due to the same segments are equally preferable. In~\Eqref{eq:CPA}, the preference labels of source task trajectory pairs are weighted by the trajectory pair correspondence. This means that the preference labels of matched trajectory pairs contribute more to the preference transfer. The full procedures for computing \CPA labels are shown in Algorithm~\ref{alg:CPA} in Appendix~\ref{app:algorithm}.

\subsection{Robust Reward Learning}

Obtaining preferences labels transferred according to the \ot matrix, we can utilize \pbRL approaches, such as the offline \pbRL algorithm PT~\citep{kim2023preference}, to learn reward functions. However, the labels may include some noise and learning from such data using previous methods will influence the accuracy of the rewards and eventually the performance of the policy.

Prior \pbRL methods represent the rewards as fixed scalar values~\citep{christiano2017deep, lee2021pebble, kim2023preference}. However, this type of reward modeling is vulnerable to noisy labels. Given a preference dataset comprising trajectory pairs and their preference labels, altering one preference label $z$ of a pair $(x_0,x_1)$ into $1-z$ will dramatically change the optimization direction of the reward function on the pair. Thus, if we respectively learn two reward models from the clean dataset and the data with an inverse label, the two reward models will predict distinct values for that trajectory pair. Subsequent, the inaccurate rewards will affect the performance of the policy. Therefore, a robust \pbRL algorithm capable of learning from noisy labels is necessary.

\paragraph{Distributional Reward Modeling.}

To improve the robustness of \pbRL in the presence of noisy labels, we incorporate reward uncertainty and model the rewards from a distributional perspective. Specifically, the rewards are modeled as Gaussian distributions, where the mean represents the estimated reward and the variance signifies the reward uncertainty. 

As shown in Figure~\ref{fig:RPT}, we design two branches for modeling reward mean and variance concurrently. Given the extracted embedding $\{\vecx_t\}_{t=1}^H$ of a trajectory segment of length $H$, we split $\{\vecx_t\}$ into two tensors of the same shape along the embedding dimension. These split tensors are separately processed by the mean and variance branches, ultimately yielding reward mean $\{\hat{r}_t\}$ and variance $\{\sigma_t^2\}$. With reward mean and variance, we then construct the preference predictor $P_\psi$ and derive the loss function for distributional reward learning based on~\Eqref{eq:sup_loss}:
\begin{equation}
\begin{aligned}
    \gL_{\text{ce}} =& \underset{{(x^0,x^1,z)\sim\gD}}{\E} \Big[\mathrm{CE}\big(P_\psi(\{\hat{r}_t^0\}, \{\hat{r}_t^1\}), z\big) \\
    &+ \lambda \cdot \E_{\beta_t^0 \sim p(\beta_t^0), \beta_t^1 \sim p(\beta_t^1)}\mathrm{CE}\big(P_\psi(\{\beta_t^0\}, \{\beta_t^1\}), z\big) \Big],
\end{aligned}
\label{eq:robust_sup_loss}
\end{equation}
where $\lambda$ balances the reward mean $\{\hat{r}_t\}$ and the stochastic term $\{\beta_t\}$, $\{\hat{r}_t^0\}$ and $\{\beta_t^0\}$ respectively denote the reward mean and reward samples of trajectory segment $x^0$ (and $\{\hat{r}_t^1\}$ and $\{\beta_t^1\}$ for $x^1$), preference predictor $P_\psi$ in the first term takes the reward mean of two segments as inputs while the second $P_\psi$ uses sampled rewards of two segments as inputs, and $\mathrm{CE}$ denotes the cross-entropy loss. In practical, the second expectation in~\Eqref{eq:robust_sup_loss} is approximated by the mean of $K$ samples from the distribution of $\beta$.

\paragraph{Regularization Loss.}

The sampled rewards with large variance will make the second term of~\Eqref{eq:robust_sup_loss} a large value. If we directly optimize~\Eqref{eq:robust_sup_loss}, the variance of all samples will decrease, and eventually close to zero.
Therefore, to avoid the variance collapse, we introduce a regularization loss to force the uncertainty level to maintain a level $\eta$:
\begin{equation}
    \gL_{\text{reg}} = \max(0, \eta - h(\gN(\hat{r}, \sigma^2))),
\label{eq:reg_loss}
\end{equation}
where $h(\gN(\hat{r}, \sigma^2)) = \frac{1}{2} \log (2\pi e\sigma^2)$ computes the entropy of the Gaussian distribution. Combing the cross-entropy loss in~\Eqref{eq:robust_sup_loss} and regularization loss in~\Eqref{eq:reg_loss}, the total loss for RPT training is as follows:
\begin{equation}
    \gL(\psi) = \gL_{\text{ce}} + \alpha \cdot \gL_{\text{reg}},
\label{eq:total_loss}
\end{equation}
where $\alpha$ is a trade-off factor between the two terms.

\paragraph{Reparameterization Trick.}

Directly sampling $\beta$ from $\gN(\hat{r}, \sigma^2)$ will make the back propagation process intractable. Hence, we use the reparameterization trick to first sample a noise $\epsilon$ from standard Gaussian distribution $\gN(0,1)$, and computes the sample by:
\begin{equation}
    \beta = \hat{r} + \sigma \cdot \epsilon.
\end{equation}
Therefore, the reward mean and variance can be learned without the influences of sampling operation.

\begin{table*}[!t]
\centering
\caption{Success rate of our method against the baselines on robotic manipulations tasks of Meta-World and Robomimic benchmark. The results are reported with mean and standard deviation across five random seeds. The results of \ourmethod are \textbf{bolded} and the last column denotes the accuracy of computing \CPA labels. The results demonstrate that our method exceeds PT+Sim and is comparable to PT with ground-truth scripted labels.}
\resizebox{1\linewidth}{!}{
\begin{tabular}{ll|rrr|rrrr|r}
    \toprule
    \multirow{2}{*}{\bf Source Task}
    & \multirow{2}{*}{\bf Target Task}
    & \multicolumn{3}{c}{\bf PbRL with Scripted Labels}
    & \multicolumn{4}{c}{\bf PbRL with Transferred Labels}
    & \multirow{2}{*}{\begin{tabular}[c]{@{}c@{}} \bf \CPA \\ \bf Acc. \end{tabular}} \\ \cmidrule(lr){3-5} \cmidrule(lr){6-9}
    &  & \multicolumn{1}{c}{\bf PT} & \multicolumn{1}{c}{\bf PT+Semi} & \multicolumn{1}{c}{\bf IPL} & \multicolumn{1}{c}{\bf PT+Sim} & \multicolumn{1}{c}{\bf PT+\CPA} & \multicolumn{1}{c}{\bf RPT+\CPA} & \multicolumn{1}{c}{\bf IPL+\CPA} & \\
    \midrule
    \multirow{4}{*}{Button Press} & Window Open
    & \mean{89.2} \std{5.4}
    & \mean{86.4} \std{3.0}
    & \mean{91.6} \std{6.2}
    & \mean{44.0} \std{26.3}
    & \bmean{85.6} \std{17.1}
    & \bmean{88.0} \std{11.6}
    & \bmean{91.2} \std{5.9}
    & \mean{87.0} \\
     & Door Close
    & \mean{94.8} \std{4.8}
    & \mean{94.8} \std{7.6}
    & \mean{75.6} \std{32.6}
    & \mean{63.6} \std{24.5}
    & \bmean{59.6} \std{49.1}
    & \bmean{78.4} \std{29.5}
    & \bmean{46.8} \std{30.7}
    & \mean{78.0} \\
     & Drawer Open
    & \mean{96.6} \std{6.1}
    & \mean{96.8} \std{3.3}
    & \mean{91.2} \std{4.1}
    & \mean{18.0} \std{33.0}
    & \bmean{80.8} \std{21.0}
    & \bmean{84.0} \std{16.0}
    & \bmean{76.8} \std{10.4}
    & \mean{76.6} \\
     & Sweep Into
    & \mean{86.0} \std{8.7}
    & \mean{88.4} \std{5.2}
    & \mean{73.2} \std{6.4}
    & \mean{48.8} \std{34.9}
    & \bmean{77.2} \std{11.0}
    & \bmean{80.0} \std{6.8}
    & \bmean{76.8} \std{7.6}
    & \mean{69.5} \\
    \midrule
    \multirow{4}{*}{Faucet Close} & Window Open
    & \mean{89.2} \std{5.4}
    & \mean{86.4} \std{3.0}
    & \mean{91.6} \std{6.2}
    & \mean{21.2} \std{17.2}
    & \bmean{84.8} \std{10.9}
    & \bmean{88.8} \std{6.7}
    & \bmean{88.4} \std{11.5}
    & \mean{87.0} \\
     & Door Close
    & \mean{94.8} \std{4.8}
    & \mean{94.8} \std{7.6}
    & \mean{75.6} \std{32.6}
    & \mean{38.8} \std{44.8}
    & \bmean{72.8} \std{40.9}
    & \bmean{86.4} \std{8.2}
    & \bmean{41.6} \std{31.5}
    & \mean{72.0} \\
     & Drawer Open
    & \mean{96.6} \std{6.1}
    & \mean{96.8} \std{3.3}
    & \mean{91.2} \std{4.1}
    & \mean{56.4} \std{23.4}
    & \bmean{79.2} \std{8.8}
    & \bmean{90.8} \std{12.0}
    & \bmean{70.4} \std{11.6}
    & \mean{77.0} \\
     & Sweep Into
    & \mean{86.0} \std{8.7}
    & \mean{88.4} \std{5.2}
    & \mean{73.2} \std{6.4}
    & \mean{14.0} \std{20.0}
    & \bmean{71.6} \std{17.4}
    & \bmean{75.2} \std{6.6}
    & \bmean{81.6} \std{7.1}
    & \mean{68.4} \\
    \midrule
    \multirow{2}{*}{Square-MH} & Can-MH    
    & \mean{35.6} \std{11.6}
    & \mean{30.8} \std{12.7}
    & \mean{50.8} \std{12.2}
    & \mean{-}
    & \bmean{32.8} \std{5.9}
    & \bmean{34.8} \std{12.1}
    & \bmean{45.6} \std{8.2}
    & \mean{70.0} \\
     & Lift-MH   
    & \mean{68.8} \std{19.2}
    & \mean{60.8} \std{7.3}
    & \mean{92.4} \std{3.3}
    & \mean{-}
    & \bmean{62.0} \std{19.1}
    & \bmean{74.4} \std{23.1}
    & \bmean{81.6} \std{6.1}
    & \mean{63.2} \\
    \midrule
    \multicolumn{2}{c|}{\bf Average (w/o Robomimic)} & \mean{91.7} & \mean{91.6} & \mean{82.9} & \mean{38.1} & \bmean{76.5} & \bmean{84.0} & \bmean{71.7} & \mean{76.9} \\
    \multicolumn{2}{c|}{\bf Average (all settings)}      & \mean{83.8} & \mean{82.4} & \mean{80.6} & \mean{-} & \bmean{71.2} & \bmean{78.1} & \bmean{70.1} & \mean{74.9} \\
    \bottomrule
\end{tabular}
}
\label{tab:main_results}
\end{table*}

\subsection{Practical Algorithm}

The entire algorithm mainly comprises three stages. First, our approach computes \CPA labels based on \GWd alignment and the procedures are shown in Algorithm~\ref{alg:CPA} in Appendix~\ref{app:algorithm}. In Algorithm~\ref{alg:CPA}, we use Sinkhorn algorithm~\citep{peyre2016gromov} to solve the \ot matrix, which is implemented by Python Optimal Transport~\citep{flamary2021pot}.
For post-processing, we apply min-max normalization to \CPA labels computed by~\Eqref{eq:CPA}, as the numerical values of the labels are relatively small after being weighted by $\matA$. Additionally, we binarize the normalized labels, transforming them from continuous values into standard discrete preference labels by~\Eqref{eq:binarize}:
\begin{equation}
    z(y_j, y_{j^\prime}) = 
    \begin{cases}
    1 & \text{if } z(y_j, y_{j^\prime}) > 0.5, \\
    0 & \text{otherwise}
    \end{cases}
\label{eq:binarize}
\end{equation}
Second, we implement the distributional reward modeling based on PT~\cite{kim2023preference} and name it Robust Preference Transformer~(RPT). RPT is trained by \CPA labels obtained from the first step. Last, we relabel the transitions in the offline dataset using the trained reward function and train offline RL algorithms, such as Implicit Q-Learning (IQL)~\citep{kostrikov2022offline}. The full procedures are shown in Algorithm~\ref{alg:main} in Appendix~\ref{app:algorithm}.

\section{Experiments}

In this section, we first conduct experiments to evaluate our proposed method on several pairs of robotic manipulation tasks from Meta-World~\citep{yu2020meta} and Robomimic~\citep{mandlekar2021matters} in zero-shot setting. Then we demonstrate our approach significantly surpasses existing methods in limited-data scenarios. Last, we evaluate our algorithm with different choice of cost functions and noise levels.

\subsection{Compared Methods and Training Details}
\label{sec:5.1}

The following methods are included for experimental evaluation:

\begin{itemize} [leftmargin=20pt]
    \setlength\itemsep{-0.1em}
    \setlength\topsep{0em}
    \item PT~\citep{kim2023preference}: The original PT algorithm trained from the preference labels computed by the ground-truth rewards.
    \item PT+Semi: This baseline combines PT with semi-supervised learning, which is proposed in the online feedback-efficient \pbRL algorithm SURF~\citep{park2022surf}. The method infers pseudo preference labels of unlabeled data based on the reward confidence.
    \item IPL~\citep{hejna2023inverse}: An offline \pbRL algorithm that learns policies without modeling reward functions.
    \item PT+Sim: This baseline is a cross-task \pbRL algorithm that calculates transferred preference labels simply based on the trajectory similarity between tasks.
    \item PT+\CPA (Ours): The method is a zero-shot \pbRL algorithm that learns PT from preference labels transferred by \CPA.
    \item RPT+\CPA (Ours): The method robustly learns RPT from \CPA labels by modeling reward distribution.
    \item IPL+\CPA (Ours): The method learns from \CPA labels without reward functions.
\end{itemize}

\paragraph{Implementation Details.}

\begin{figure*}[t]
\centering
\begin{tabular}{ccc}
\subfloat[Door Close]{\includegraphics[width=0.28\linewidth]{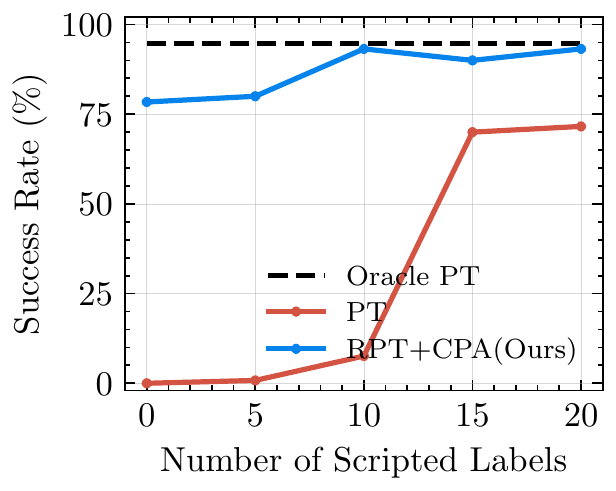}}
& \subfloat[Window Open]{\includegraphics[width=0.28\linewidth]{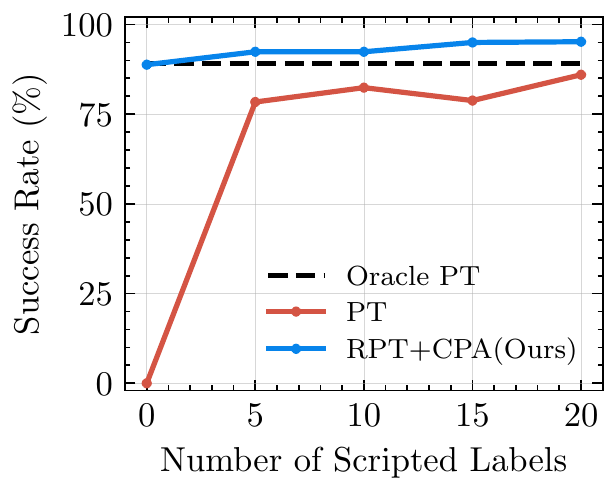}}
& \subfloat[Lift-MH]{\includegraphics[width=0.28\linewidth]{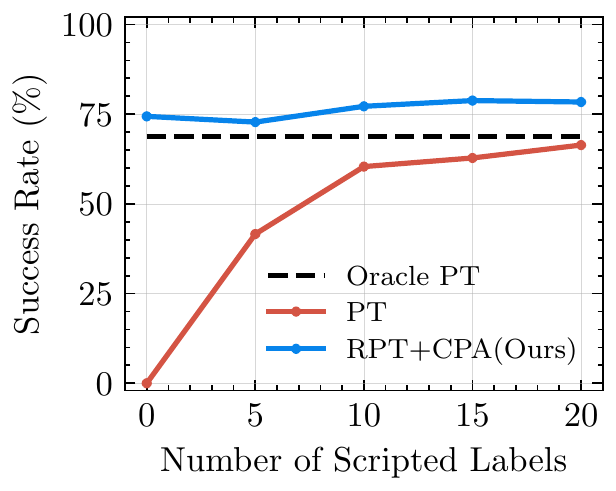}}
\end{tabular}
\caption{Success rate of Door Close, Window Open and Lift-MH with different scripted preference labels.}
\label{fig:abla_same_oracle}
\end{figure*}

\begin{table*}[!ht]
\centering
\caption{Success rates on three pairs of source and target tasks with different cost functions. The results are reported with mean and standard deviation of success rate across five runs.}
\resizebox{0.65\linewidth}{!}{
\begin{tabular}{llrrrrr}
    \toprule
    \multirow{2}{*}{\bf Source Task} & \multirow{2}{*}{\bf Target Task} & \multicolumn{2}{c}{\bf Euclidean} & \multicolumn{2}{c}{\bf Cosine} \\
    \cmidrule(l){3-4}\cmidrule(l){5-6}
    & & \multicolumn{1}{c}{\bf RPT+\CPA} & \multicolumn{1}{c}{\bf \CPA Acc.} & \multicolumn{1}{c}{\bf RPT+\CPA} & \multicolumn{1}{c}{\bf \CPA Acc.} \\
    \midrule
    Button Press & Sweep Into  
    & \mean{80.0} \std{6.8}
    & \mean{69.5}
    & \mean{79.2} \std{5.4}
    & \mean{65.0} \\
    Faucet Close & Window Open 
    & \mean{88.8} \std{6.7}
    & \mean{87.0}
    & \mean{92.4} \std{3.6}
    & \mean{91.0} \\
    Square-MH    & Lift-MH 
    & \mean{74.4} \std{23.1}
    & \mean{63.2}
    & \mean{69.3} \std{9.5}
    & \mean{66.0} \\
    \midrule
    \multicolumn{2}{c}{\bf Average} & \mean{81.1} & \mean{73.2} & \mean{80.3} & \mean{74.0} \\
    \bottomrule
\end{tabular}
}
\label{tab:abla_cost}
\end{table*}

All methods are implemented based on the officially released code of PT~\footnote{\href{https://github.com/csmile-1006/PreferenceTransformer}{https://github.com/csmile-1006/PreferenceTransformer}} and IPL~\footnote{\href{https://github.com/jhejna/inverse-preference-learning}{https://github.com/jhejna/inverse-preference-learning}}. RPT is implemented by replacing the preference attention layer of PT with two branches, each comprising a two-layer Multi-layer Perceptrons~(MLPs), with the other settings identical to PT. Both PT and PT+Semi utilize scripted labels computed according to ground-truth rewards, which is a common way for the evaluation of \pbRL algorithms~\citep{lee2021pebble, liu2022meta, kim2023preference}. PT+\CPA, RPT+\CPA and IPL+\CPA are trained with computed \CPA labels (zero-shot) or a mixture of \CPA labels and scripted labels (few-shot). All PT-based methods initially train reward models using the preference data, and the offline RL algorithm IQL~\citep{kostrikov2022offline} is used for policy learning following PT. For IPL-based method, policies are directly learned from preferences.

For the Meta-World benchmark, Button Press and Faucet Close serve as source tasks, while Window Open, Door Close, Drawer Open, and Sweep Into are evaluated as target tasks. For Robomimic, we set Square-MH as the source task, and Can-MH and Lift-MH as target tasks. The used tasks and datasets are detailed in Appendix~\ref{app:exp_details}.
We set the segment length as $50$ for Meta-World tasks and $100$ for Robomimic tasks. For the number of target task preference labels, we provide $100$ for Window Open and Door Close, $500$ for Drawer Open, Can-MH and Lift-MH, and $1000$ for Sweep Into. The Euclidean function is employed as the cost function in the \GWd alignment, with different cost functions discussed in Section~\ref{sec:5.3}. Regarding RPT learning, the margin $\eta$ in~\Eqref{eq:reg_loss} is set to $100$ for all experiments, with different margin effects evaluated in Appendix~\ref{app:results}. The number of samples $K$ in~\Eqref{eq:robust_sup_loss} is consistently set to $5$. The weight $\lambda$ in~\Eqref{eq:robust_sup_loss} is $0.3$ for Door Close with Button Press as the source task, and $0.1$ for the other task pairs. The trade-off $\alpha$ in~\Eqref{eq:total_loss} is set to $0.02$ for Drawer Open with Button Press as the source task, and $0.01$ for all other experiments. Detailed network architectures and hyperparameters of all methods and IQL are presented in Appendix~\ref{app:exp_details}.

The tasks of Meta-World and Robomimic are evaluated through success rate. Each task is conducted with five random seeds, with the mean and standard deviation of success rates reported. Each run evaluates the policy by rolling out $50$ episodes at every evaluation step, calculating performance as the mean success rate over these $50$ episodes. All experiments are run on NVIDIA GeForce RTX 3080 and NVIDIA Tesla V100 GPUs with 8 CPU cores.

\subsection{Results of Zero-shot Preference Learning}
\label{sec:5.2}

Table~\ref{tab:main_results} shows the results on robotic manipulation tasks of Meta-World and Robomimic with different pairs of source and target tasks~\footnote{PT, PT+Semi and IPL do not require preference data from source tasks, so the their results are solely depend on the target tasks.}\footnote{PT+Sim cannot work by transferring preferences from Square-MH to Lift-MH because the state dimension of these two tasks are different.}. For the baselines that use scripted preference labels, PT, PT+Semi and IPL yield outstanding performance on the majority of tasks, where PT achieves a mean success rate of $91.7\%$ on Meta-World Tasks and $83.8\%$ across all tasks. The performance of PT+Semi is almost the same with that of PT on Meta-World tasks, but has a drop on Robomimic tasks. For IPL, it outperforms PT and PT+Semi on Robomimic, while its performance on Meta-World is worse than that of them. By transferring preference via OT, \CPA attains a mean accuracy of $74.9\%$ in computing preference labels across all tasks. PT trained with \CPA labels realizes a $71.2\%$ success rate, equating to $85.0\%$ of oracle performance (i.e., the performance of PT trained with scripted labels). RPT, incorporating reward uncertainty in reward modeling, enhances performance to $78.1\%$ across all tasks when trained with \CPA labels, equivalent to \bmean{93.2\%} oracle performance. Also, IPL+\CPA achieves $70.1\%$ success rate across all tasks, which is equal to $87.0\%$ oracle performance. We can conclude that RPT+\CPA can achieve competitive performance compared with baselines trained with scripted preference labels. Both PT+\CPA and RPT+\CPA outperforms PT+Sim by a large margin, and RPT+\CPA even exceeds PT on the Lift-MH task, demonstrating the powerful capabilities of \CPA and RPT in zero-shot preference transfer and robust learning.

\begin{figure*}[t]
\centering
\begin{tabular}{ccc}
\subfloat[Sweep Into]{\includegraphics[width=0.28\linewidth]{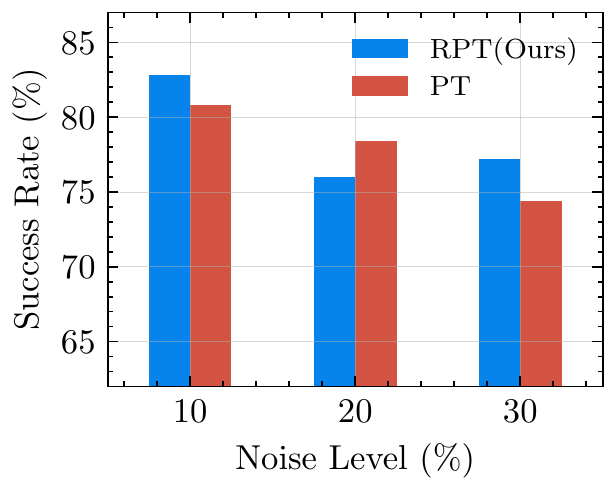}}
& \subfloat[Window Open]{\includegraphics[width=0.28\linewidth]{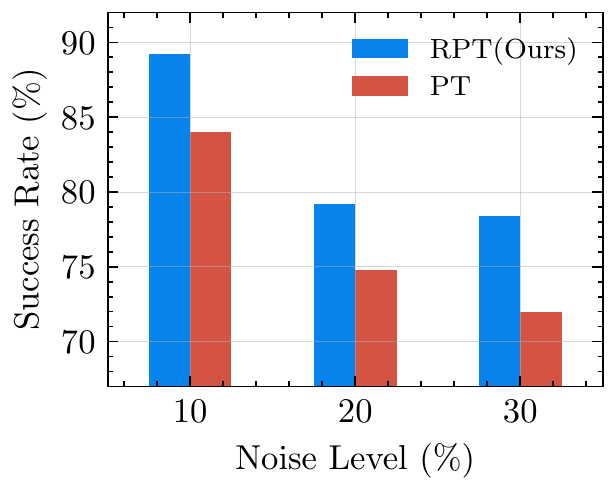}}
& \subfloat[Lift-MH]{\includegraphics[width=0.28\linewidth]{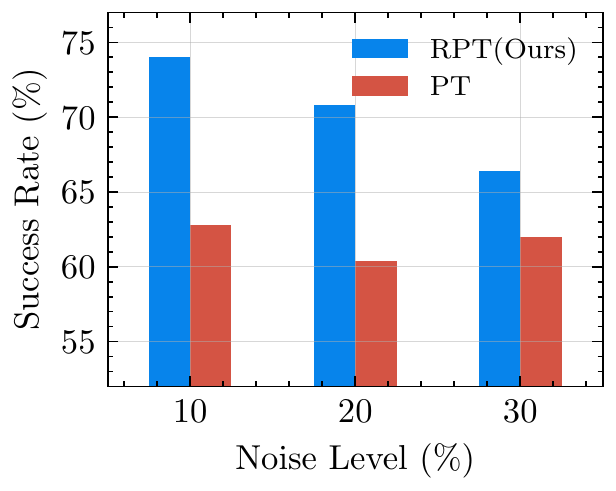}}
\end{tabular}
\caption{Success rate of Sweep Into, Window Open and Lift-MH under different noise levels.}
\label{fig:abla_noise}
\end{figure*}

\subsection{Results of Few-shot Preference Learning}

The results in Table~\ref{tab:main_results} have shown strong zero-shot transfer ability of \CPA. To further balance the human labeling cost and algorithm performance, we are interested in how well does RPT+\CPA perform when there are a small number of preference labels.
For fair comparison, we evaluate our method and PT with the same number of scripted preference labels of the target task, across $\oral \in \{0, 5, 10, 15, 20\}$. Our method additionally obtains \CPA labels with the size of $\cpal = 100 - \oral$ by transferring from the source task. The results in Figure~\ref{fig:abla_same_oracle} show that RPT+\CPA significantly outperforms PT when lacking oracle preference labeling, and the advantage becomes more obvious when the number of labels is smaller. Moreover, RPT+\CPA even exceeds the Oracle PT~(i.e., PT with $100$ scripted labels) on Window Open task when $\oral \in \{5, 10, 15, 20\}$, and Lift-MH task when $\oral \in \{0, 5, 10, 15, 20\}$. The results demonstrate the excellent performance of our method when oracle labels are hard to obtain, and \CPA can be used to significantly to reduce extensive human labeling.

\subsection{Ablation Study}
\label{sec:5.3}

\paragraph{Different Cost Functions.}

The sensitivity of \CPA to the cost function is examined by evaluating PT+\CPA and RPT+\CPA performance with varying cost functions, including the Euclidean and Cosine functions.
Table~\ref{tab:abla_cost} demonstrates that \CPA performs robustly with either cost function. Notably, \CPA with the Cosine function even attains $91.0\%$ accuracy in computing \CPA labels on the Window Open task, with its success rate ($92.4\%$) surpassing PT with scripted labels on this task ($89.2\%$).

\paragraph{Different Noise Levels.}

To evaluate the performance of PT and RPT under different noise levels, we conduct experiments with $10\%, 20\%, 30\%$ noisy labels induced by flipping scripted labels. The results in Figure~\ref{fig:abla_noise} reveal the enhanced robustness of RPT to label noise, with RPT significantly outperforming PT at higher noise levels.

\section{Conclusion}

In this paper, we present \ourmethod, a novel zero-shot cross-task \pbRL algorithm, which leverages \GWd for aligning trajectory distributions across different tasks and transfers preference labels through optimal transport matrix. CPA only needs small amount of preference data from prior tasks, eliminating the need for a substantial amount pre-collected preference data or extensive human queries. Furthermore, we propose RRL, which models reward uncertainty rather than scalar rewards to learn reward models robustly. Empirical results on various robotic manipulation tasks of Meta-World and Robomimic demonstrate the effectiveness of our method in zero-shot transferring accurate preference labels and improves the robustness of learning from noisy labels. Additionally, our method significantly surpasses the current method when there are a few preference labels. By minimizing human labeling costs to a great extent, \ourmethod paves the way for the practical applications of \pbRL algorithms.

\paragraph{Limitations.}

Our method does present certain limitations. Firstly, our method is not well-suited for high-dimensional inputs due to the potential for slower processing speeds when working with high-dimensional inputs in optimal transport. Secondly, the efficiency of our algorithm relies on the same action space between source and target tasks. Therefore, our method is not suitable for the tasks like those have completely different state and action spaces. A potential solution may be utilizing representation learning methods to obtain trajectory representations and using Gromov-Wasserstein distance to align in the representation space~\citep{chen2020graph, li2022gromov}. Please refer to Appendix~\ref{app:discuss_when} for detailed discussion on how the task similarity influences \ourmethod. We recognize these limitations and view the mitigation of these issues as important directions for future exploration and development.

\section*{Acknowledgements}

This work was supported by the STI 2030-Major Projects under Grant 2021ZD0201404.

\section*{Impact Statement}

This paper presents work whose goal is to advance the field of Machine Learning. There are many potential societal consequences of our work, none of which we feel must be specifically highlighted here.

\bibliography{ref}
\bibliographystyle{icml2024}

\newpage
\appendix
\onecolumn

\section*{Appendix}

\section{Algorithm}
\label{app:algorithm}

The algorithm of computing \CPA labels and the full algorithm of our approach is shown in Algorithm~\ref{alg:CPA} and Algorithm~\ref{alg:main}, respectively.

\begin{algorithm}[!ht]
\caption{Computing \CPA Labels}
\label{alg:CPA}
\begin{algorithmic}[1]
    \REQUIRE source trajectory set $\{x_i\}_{i=1}^M$, target trajectory set $\{y_j\}_{j=1}^N$, regularization parameter $\omega$
    \STATE Initialize $\matT \gets \frac{1}{MN} \mathbf{1}_{M} \mathbf{1}_{N}^\top$, $\vecp \gets \frac{1}{M} \mathbf{1}_{M}$, $\vecq \gets \frac{1}{N} \mathbf{1}_{N}$
    \STATE Compute $\matC_\src$, $\matC_\tgt$ with $[\matC_\src]_{ij}=\left| x_i - x_j \right|_2$ and $[\matC_\tgt]_{ij}=\left| y_i - y_j \right|_2$
    \STATE Compute $\matC_{\src\tgt}$ with $\matC_{\src\tgt} \gets \matC_\src^2 \vecp \mathbf{1}_{\tgtT}^{\top} + \mathbf{1}_{\srcT} \vecq^\top (\matC_\tgt^2)^{\top}$
    \FOR{each step}
        \STATE Compute $\matC \gets \matC_{\src\tgt} - 2\matC_\src \matT (\matC_\tgt)^\top$
        \STATE Set $\vecu \gets \frac{1}{M} \mathbf{1}_{M}$, $\vecv \gets \frac{1}{N} \mathbf{1}_{N}$, $\matK \gets \exp(-\matC / \omega)$
        \FOR{$k = 1, 2, \cdots$}
            \STATE $\vecu \gets \frac{\vecp}{\matK \vecv}$, $\vecv \gets \frac{\vecq}{\matK^\top \vecu}$
        \ENDFOR
        \STATE $\matT \gets \operatorname{diag}(\vecu) \matK \operatorname{diag}(\vecv)$
    \ENDFOR
    \FOR{each $j$}
        \FOR{each $j^\prime \neq j$}
            \STATE Compute trajectory pair matching matrix $\matA$ with~\Eqref{eq:matA}
            \STATE Compute transferred preference label of $(y_j, y_{j^\prime})$ with~\Eqref{eq:CPA}
        \ENDFOR
    \ENDFOR
    \STATE Post-process \CPA labels by min-max normalization and binaryzation
    \ENSURE \CPA labels
\end{algorithmic}
\end{algorithm}

\begin{algorithm}[!ht]
\caption{\ourmethod Algorithm}
\label{alg:main}
\begin{algorithmic}[1]
    \REQUIRE Source task preference dataset $\gD_\src$, target task dataset $\gB$, reward model learning rate of $\rho$, robust term's weight coefficient $\lambda$, regularization weight coefficient $\alpha$, RPT margin $\eta$, number of reward samples $K$
    \STATE Initialize reward model $\widehat{r}_\psi$, policy $\pi_\phi$, preference dataset of target task $\gD_\tgt \gets \emptyset$
    \STATE Perform K-means clustering and group the trajectories of the target dataset into $2$ clusters
    \FOR{each step}
        \STATE Sample $\frac{N}{2}$ trajectories from each cluster within $\gB$
        \STATE Compute \CPA labels with Algorithm~\ref{alg:CPA} and store $\gD_\tgt \gets \gD_\tgt \cup \{(y_j, y_{j^\prime}, z_j)\}_{j=1}^N$
    \ENDFOR
    \FOR{each gradient step}
        \STATE Sample minibatch preference data from $\gD_\tgt$
        \STATE Sample $K$ rewards from the reward distribution computed by the outputs of RPT
        \STATE Update $\psi$ using~\Eqref{eq:total_loss} with learning rate $\rho$
    \ENDFOR
    \STATE Label rewards of the transitions in $\gB$ using trained $\widehat{r}_\psi$
    \FOR{each gradient step}
        \STATE Sample minibatch transitions from $\gB$
        \STATE Update policy $\pi_\phi$ through offline RL algorithms
    \ENDFOR
    \ENSURE policy $\pi_\phi$
\end{algorithmic}
\end{algorithm}

\section{\PT}
\label{app:PT}

\PT~\citep{kim2023preference} applies Transformer architecture to model non-Markovian rewards. For a pair of trajectory segments $(x_0, x_1)$, the non-Markovian preference predictor is given by:
\begin{equation}
    P_\psi[x^0 \succ x^1] = \frac{\exp \big(\sum_t w_t^0 \cdot \hat{r}_\psi(\{(\mathbf{s}_i^0, \mathbf{a}_i^0)\}_{i=1}^t)\big)}{\exp \big(\sum_t w_t^0 \cdot \hat{r}_\psi(\{(\mathbf{s}_i^0, \mathbf{a}_i^0)\}_{i=1}^t)\big) + \exp \big(\sum_t w_t^1 \cdot \hat{r}_\psi(\{(\mathbf{s}_i^1, \mathbf{a}_i^1)\}_{i=1}^t)\big)},
\end{equation}
where $w_t^j = w(\{(\mathbf{s}_i^j, \mathbf{a}_i^j)\}_{i=1}^H)_t, j \in \{0, 1\}$ represents the importance weight.
PT introduces a preference attention layer that models the weighted sum of rewards using the self-attention mechanism. Assume the trajectory embedding is $\vecx_t$, $\vecx_t$ is projected into a key $\veck_t$, query $\vecq_t$ and value $\hat{r}_t$. The output $z_i$ of self-attention is calculated as:
\begin{equation}
    z_i=\sum_{t=1}^H \operatorname{softmax}(\left\{\left\langle\mathbf{q}_i, \mathbf{k}_{t^{\prime}}\right\rangle\right\}_{t^{\prime}=1}^H)_t \cdot \hat{r}_t.
\end{equation}
The weighted sum of non-Markovian rewards is computed as:
\begin{equation}
    \frac{1}{H} \sum_{i=1}^H z_i=\frac{1}{H} \sum_{i=1}^H \sum_{t=1}^H \operatorname{softmax}(\left\{\left\langle\mathbf{q}_i, \mathbf{k}_{t^{\prime}}\right\rangle\right\}_{t^{\prime}=1}^H)_t \cdot \hat{r}_t=\sum_{t=1}^H w_t \hat{r}_t.
\end{equation}
Obtaining a dataset containing $\gD = \{(x^0, x^1, z)\}$, the parameters of PT can be optimized by~\Eqref{eq:sup_loss}.

\section{Experimental Details}
\label{app:exp_details}

\subsection{Tasks}
\label{app:tasks}

The used tasks are shown in Figure~\ref{fig:tasks} and the task descriptions are listed as follows:

\begin{figure*}[!ht]
\centering
\begin{tabular}{ccc}
\subfloat[Button Press]{\includegraphics[width=0.22\linewidth]{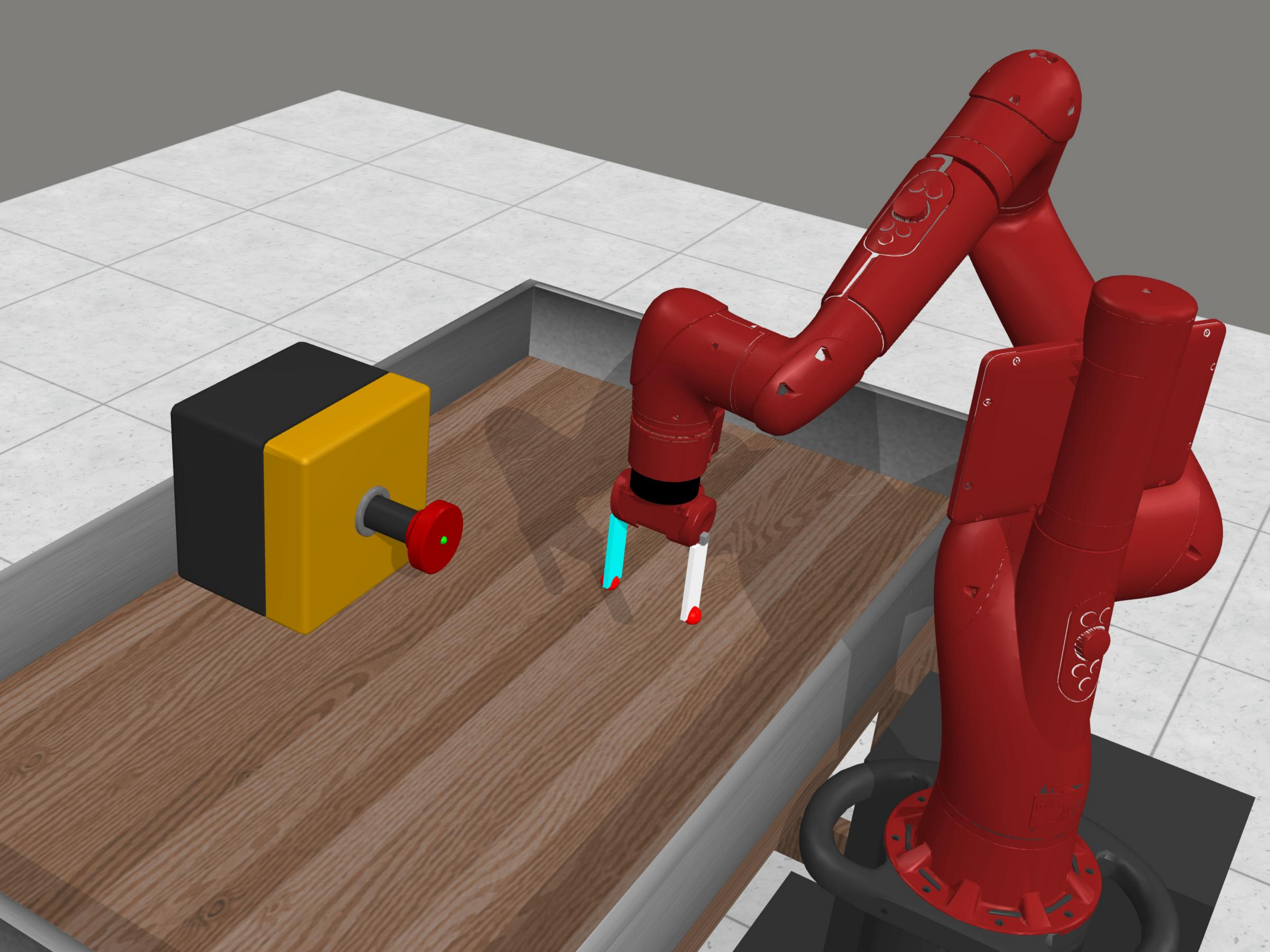}} \hspace{0.2em}
& \subfloat[Window Open]{\includegraphics[width=0.22\linewidth]{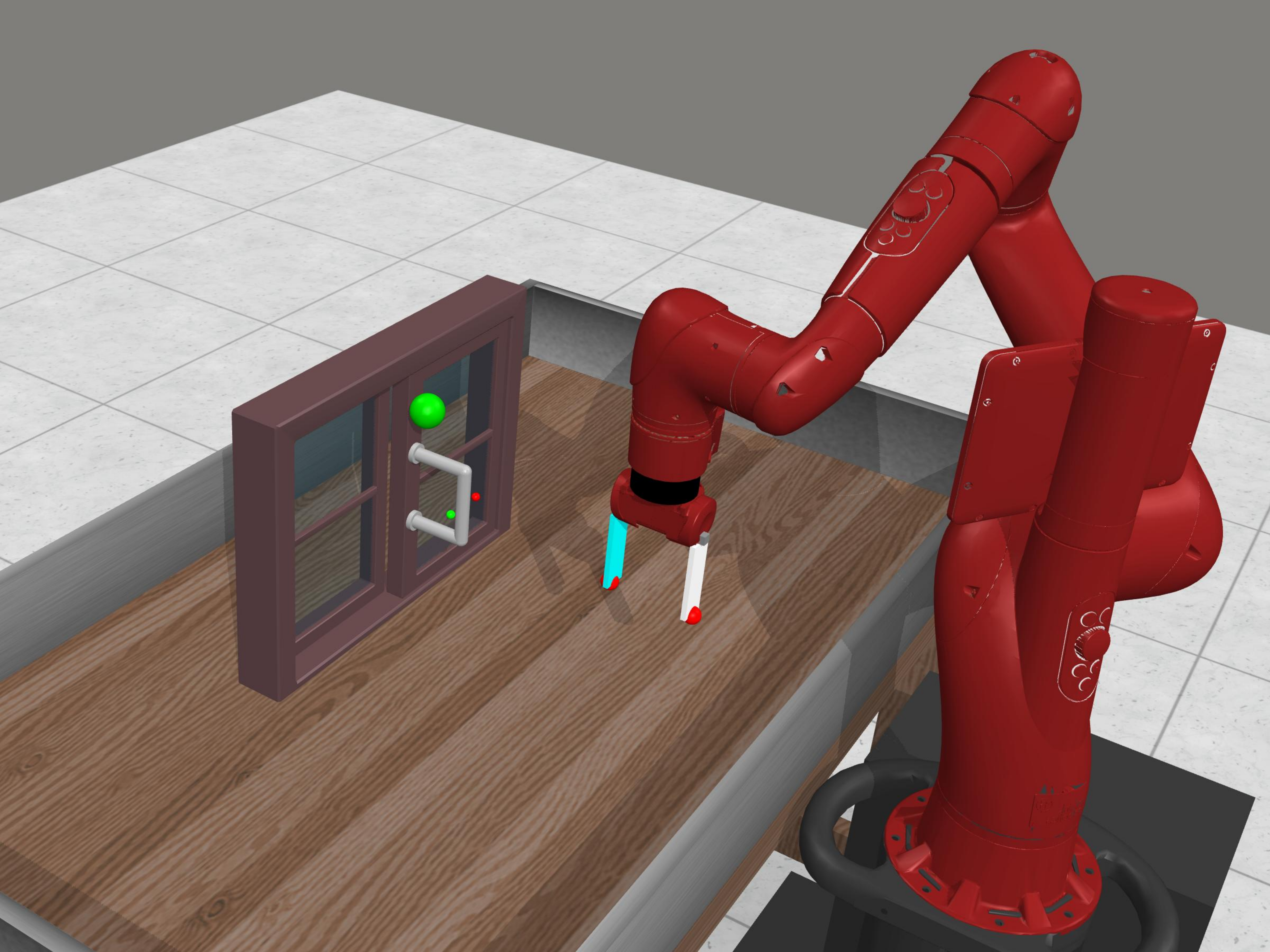}} \hspace{0.2em}
& \subfloat[Drawer Open]{\includegraphics[width=0.22\linewidth]{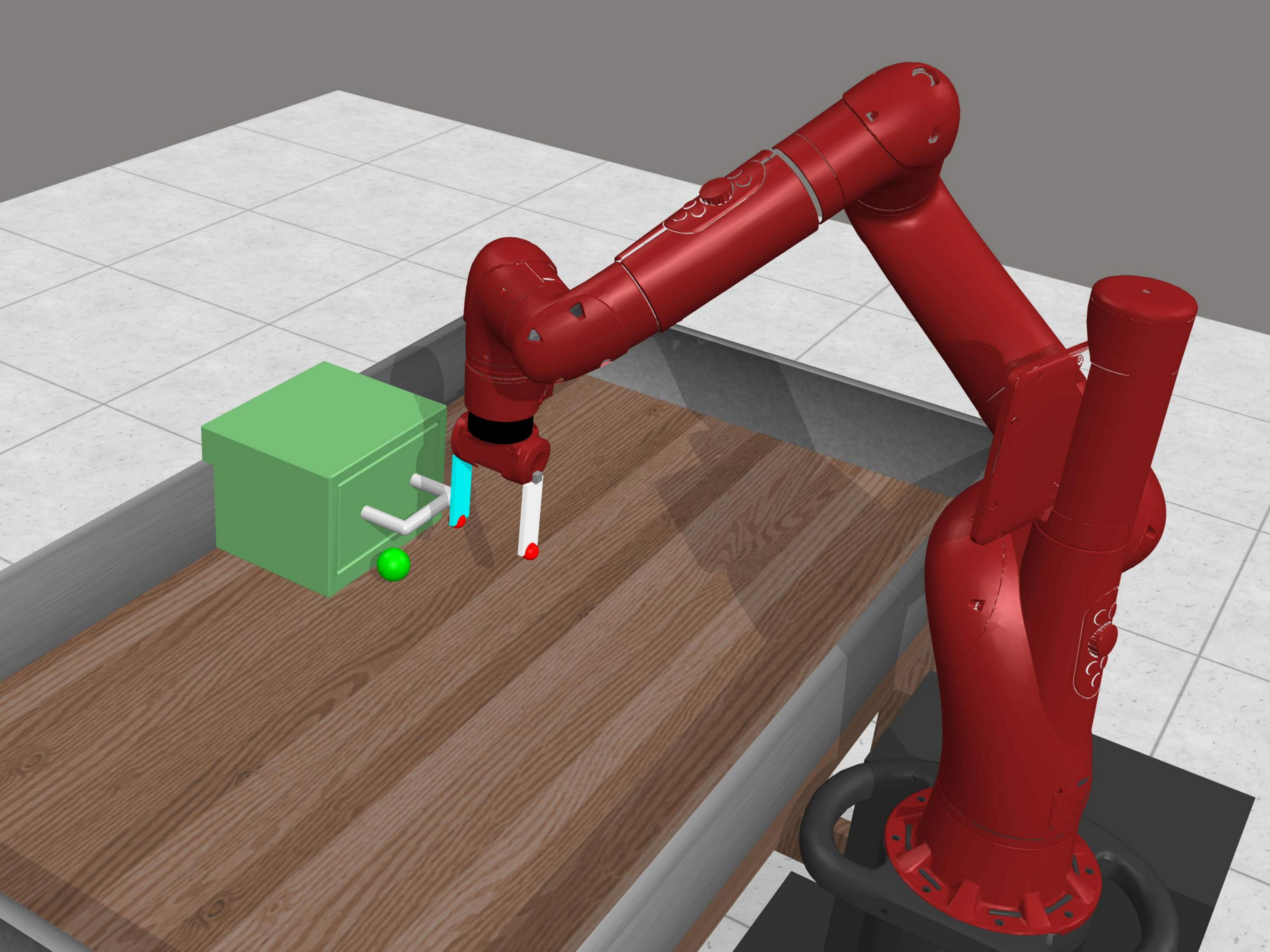}} \\
\subfloat[Faucet Close]{\includegraphics[width=0.22\linewidth]{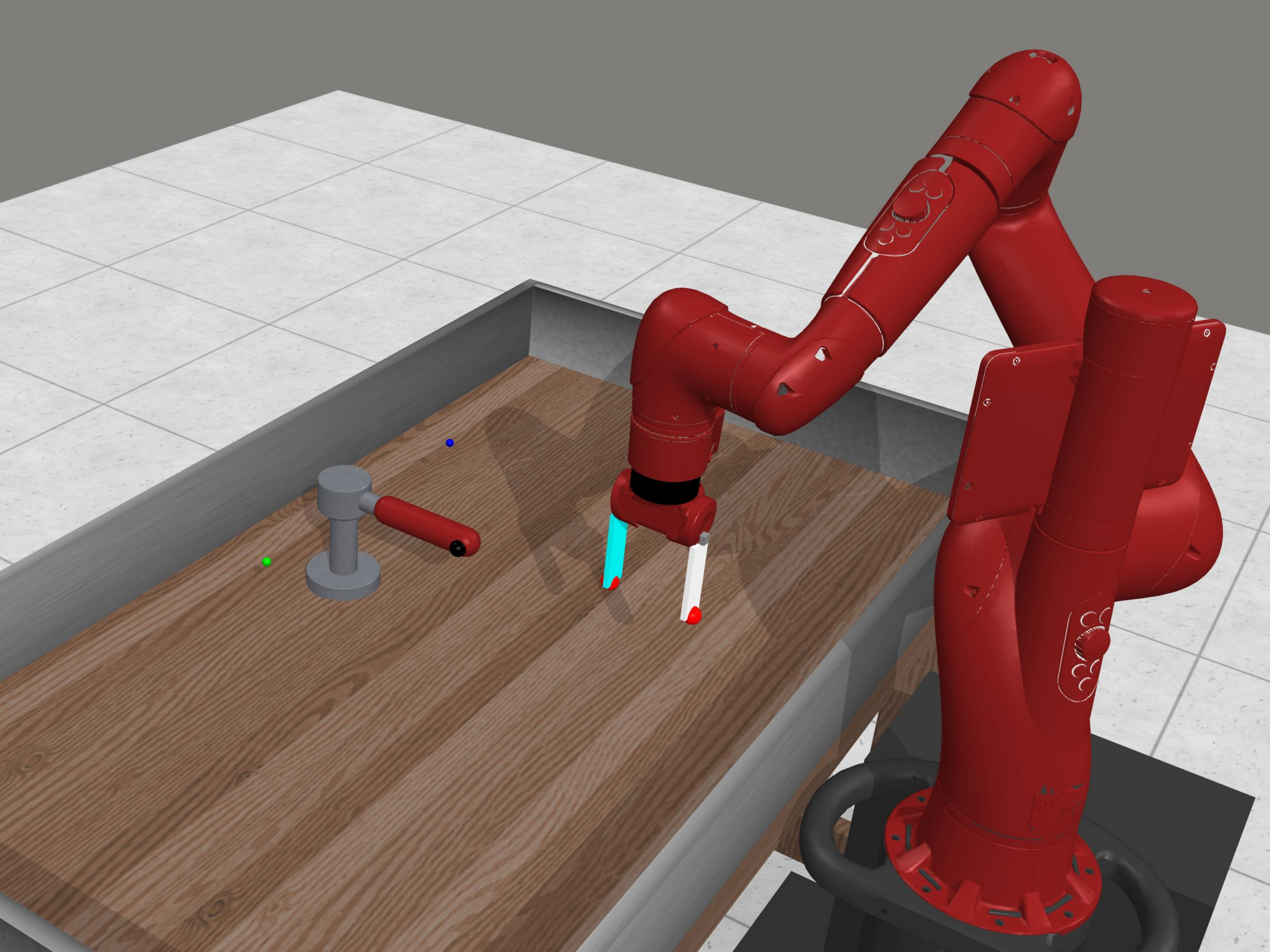}} \hspace{0.2em}
& \subfloat[Door Close]{\includegraphics[width=0.22\linewidth]{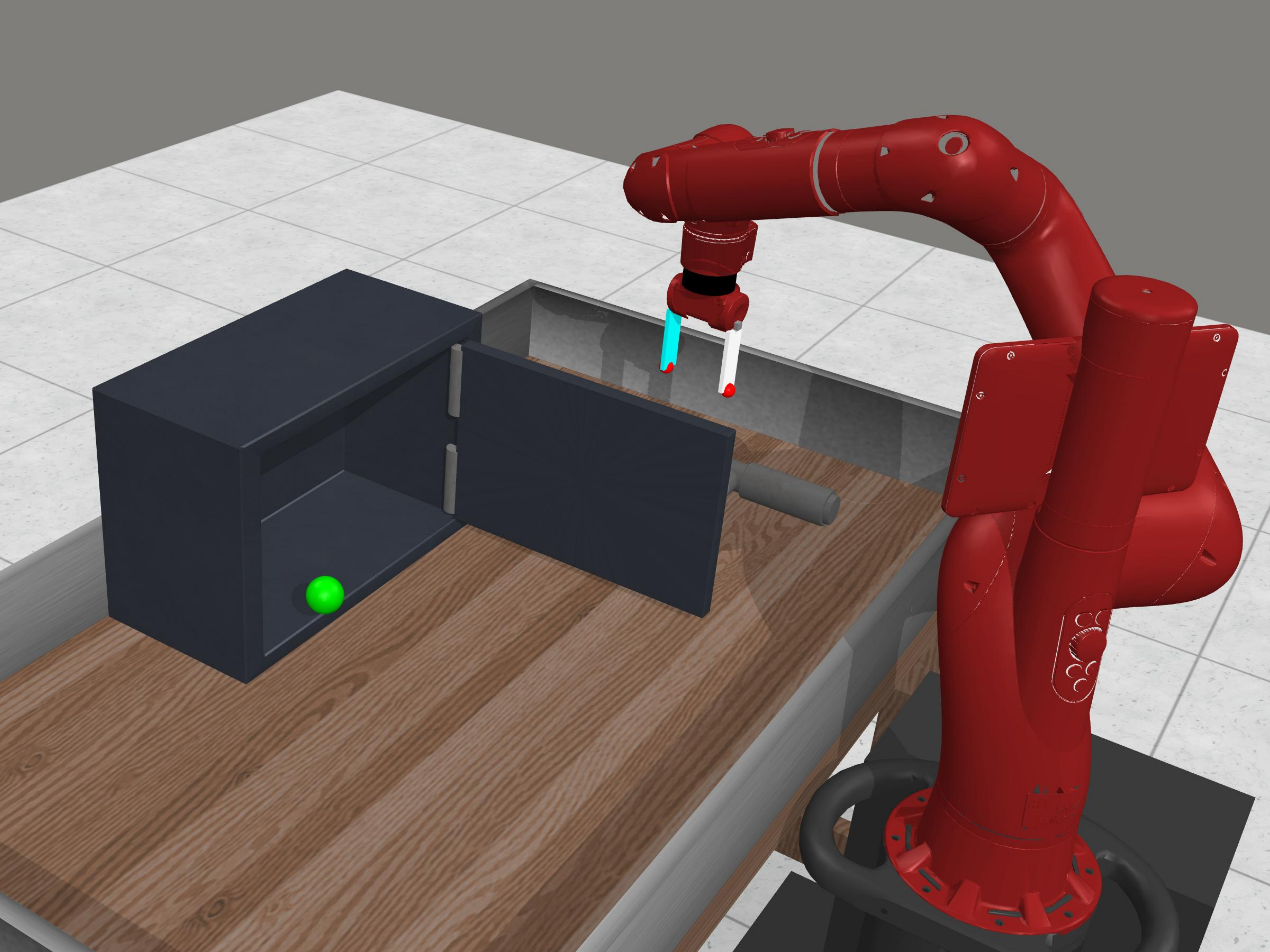}} \hspace{0.2em}
& \subfloat[Sweep Into]{\includegraphics[width=0.22\linewidth]{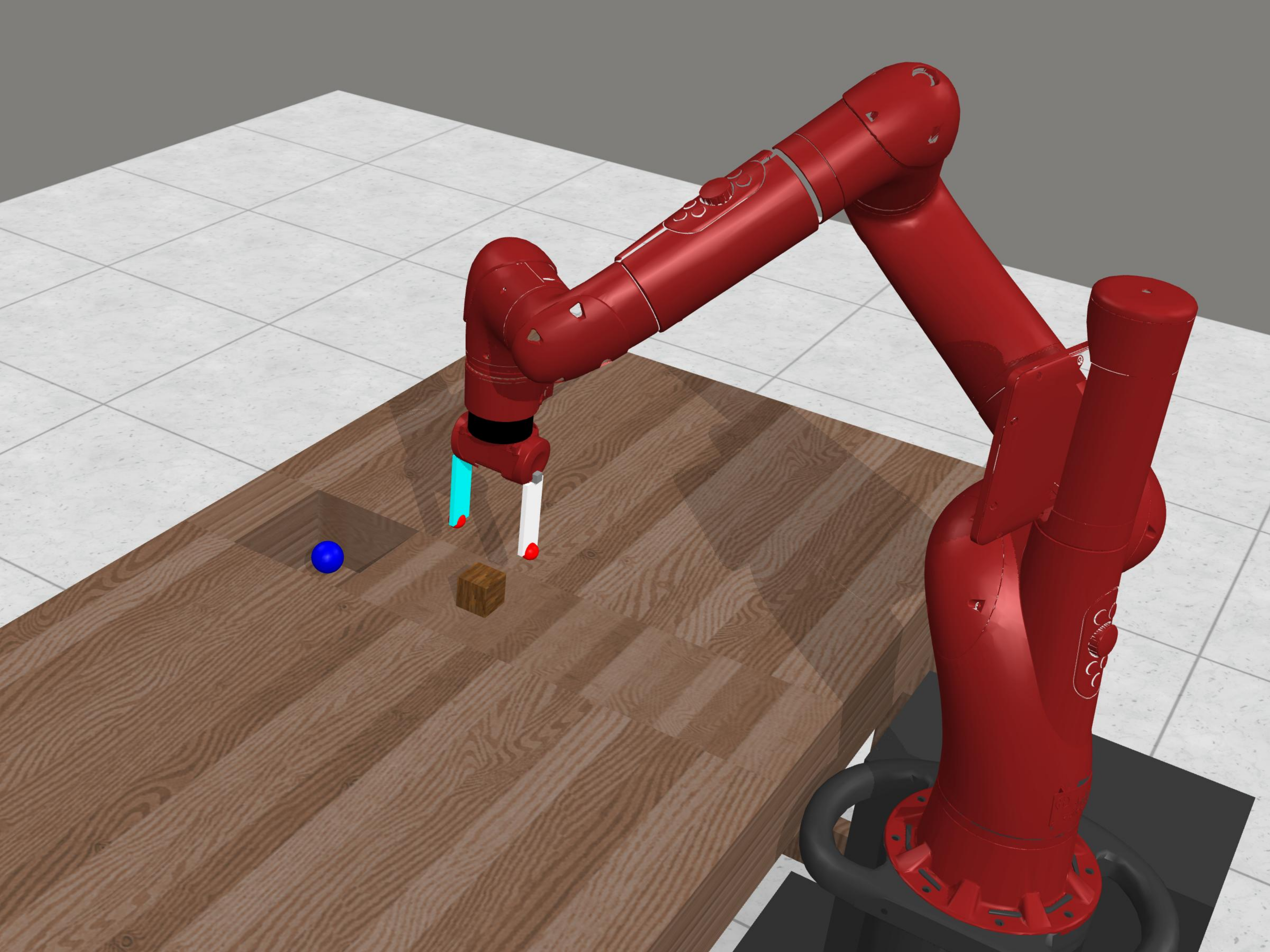}} \\
\subfloat[Square]{\includegraphics[width=0.22\linewidth]{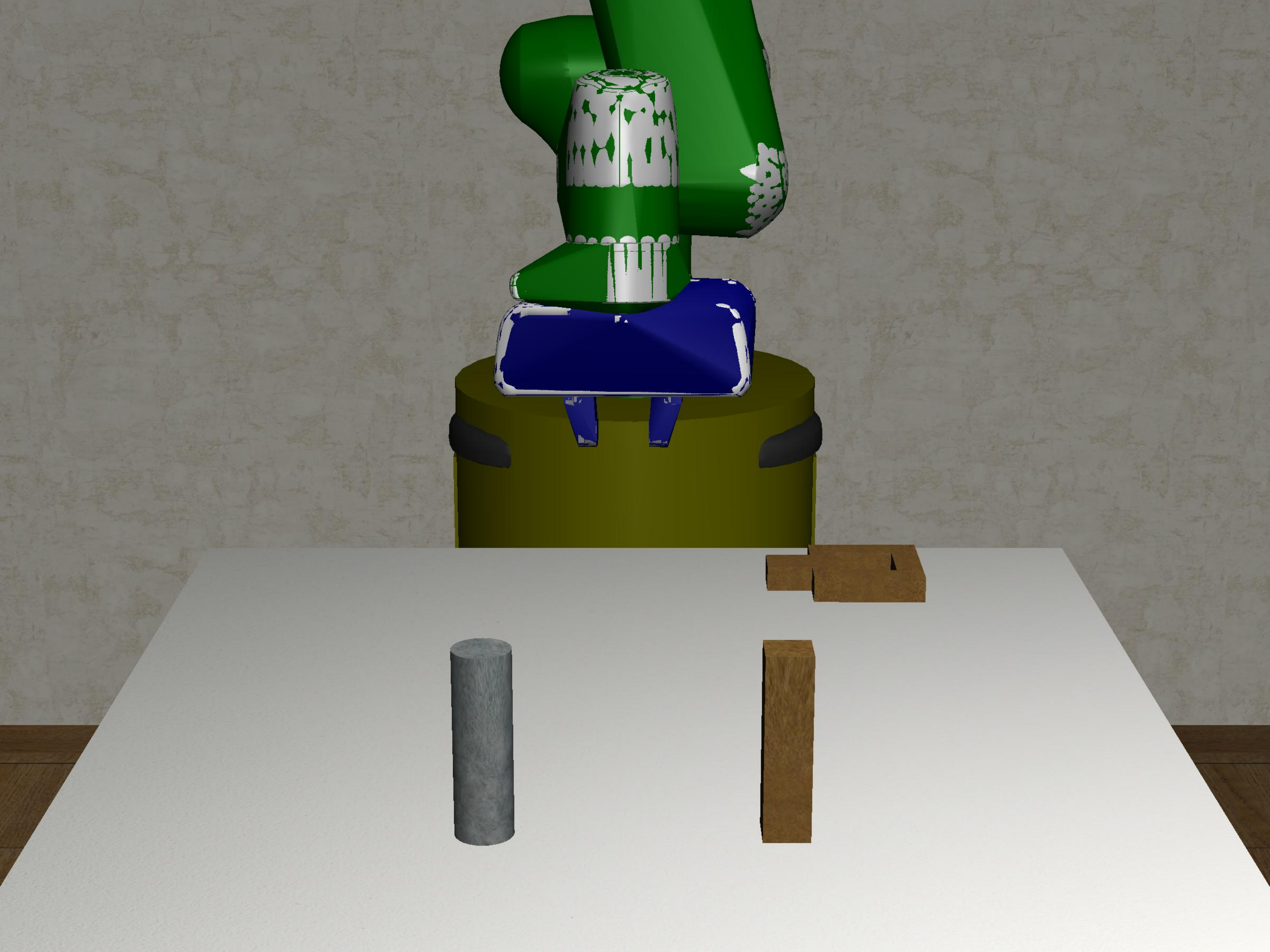}} \hspace{0.2em}
& \subfloat[Lift]{\includegraphics[width=0.22\linewidth]{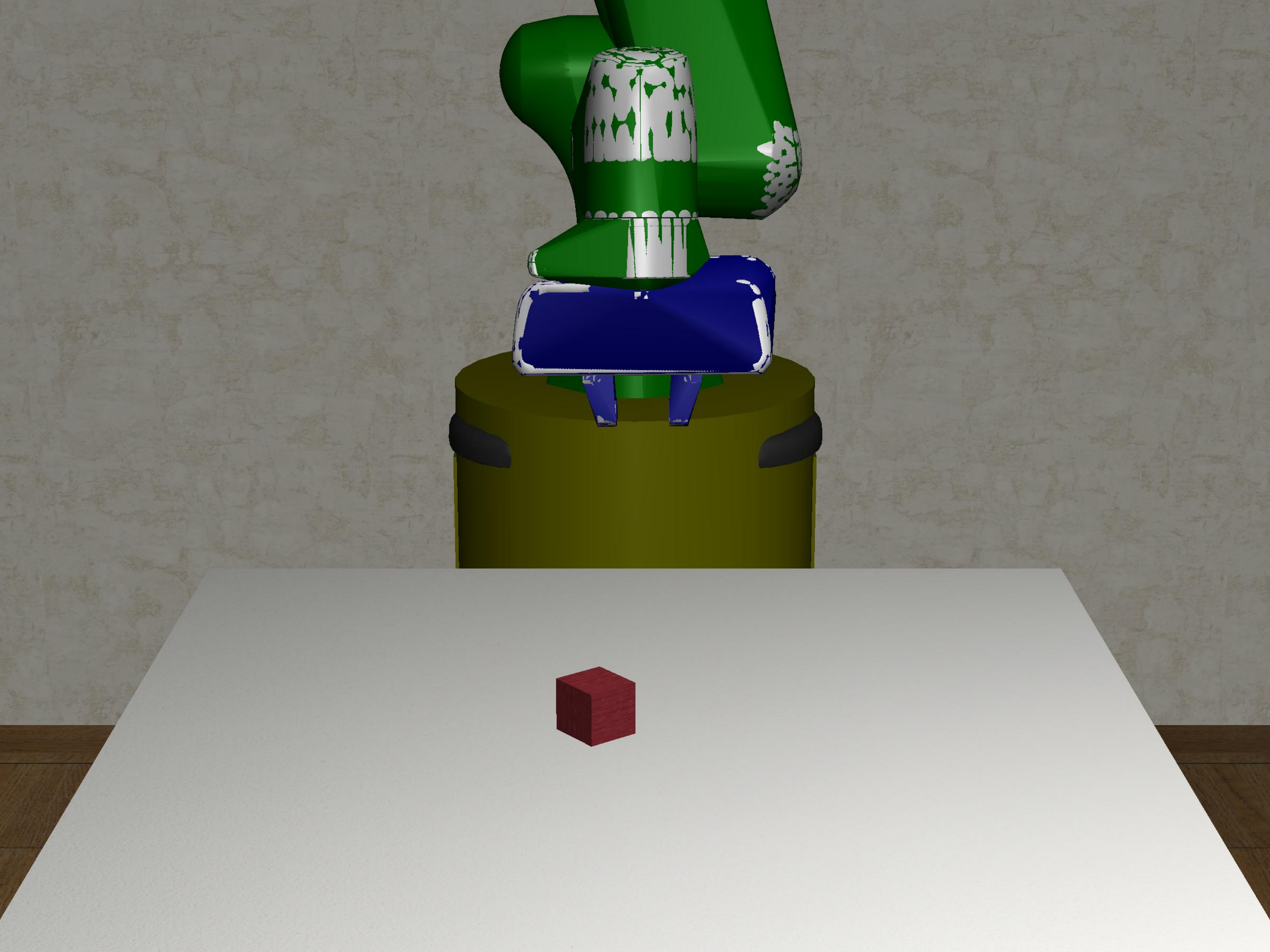}} \hspace{0.2em}
& \subfloat[Can]{\includegraphics[width=0.22\linewidth]{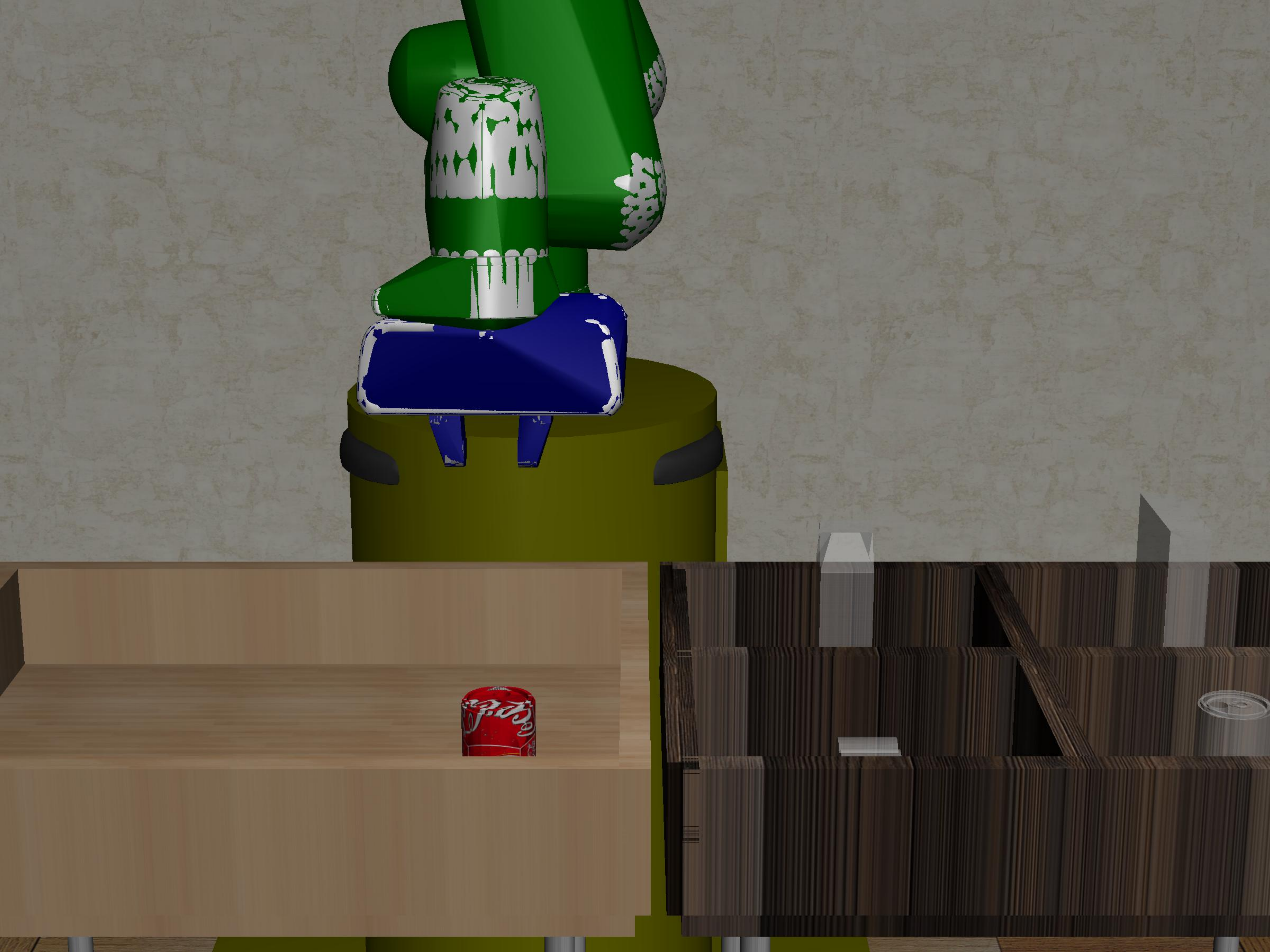}}
\end{tabular}
\caption{Nine robotic manipulation tasks used for experiments. (a-f) Meta-World tasks. (g-i) Robomimic tasks.}
\label{fig:tasks}
\end{figure*}

\paragraph{Meta-World.}

\begin{itemize} [leftmargin=20pt]
    \item Button Press: The objective is to manipulate a robotic arm to press a button. The button's initial position is arbitrarily placed.
    \item Faucet Close: The goal is to control a robotic arm to close a faucet. The initial faucet location is randomly assigned.
    \item Window Open: The task entails commanding a robotic arm to open a window. The window's starting position is randomly chosen.
    \item Door Close: The task involves guiding a robotic arm to close a door. The door's starting position is selected randomly.
    \item Drawer Open: The objective is to operate a robotic arm to open a drawer. The drawer's initial placement is arbitrary.
    \item Sweep Into: The task involves manipulating a robotic arm to propel a ball into a cavity. The ball's initial position is random.
\end{itemize}

\paragraph{Robomimic.}

\begin{itemize} [leftmargin=20pt]
    \item Square: The goal is to manipulate a robotic arm to lift a square nut and position it on a rod.
    \item Lift: The task is to operate a robotic arm to elevate a cube to a predefined height.
    \item Can: The objective is to guide a robotic arm to reposition a can from one container to another.
\end{itemize}

\subsection{Datasets}
\label{app:datasets}

\paragraph{Meta-World.}

For Meta-World tasks, the source preference datasets are collected by ground-truth policies and random policies, with the number of trajectories $M=4$. For each task, both a ground-truth policy and a random policy are utilized to roll out and obtain $2$ trajectories of length $50$.

To generate offline dataset for target tasks, we collect the replay buffer and feedback buffer using the \pbRL algorithm PEBBLE~\citep{lee2021pebble}.
For Window Open and Door Close tasks, PEBBLE is run with $120000$ steps and $2000$ scripted preference labels. For Drawer Open, we run PEBBLE with $400000$ steps and $4000$ scripted labels, and for Sweep Into, PEBBLE is run with $400000$ steps and $8000$ scripted labels.

\paragraph{Robomimic.}

The source dataset of Robomimic tasks is obtained from the Multi-Human (MH) offline dataset of each task, with the number of trajectories $M=4$. The MH dataset is collected by $6$ operators across $3$ proficiency levels, with each level comprising $2$ operators. Each operator collect $50$ demonstrations, resulting in a total of $300$ demonstrations. For each task, the source dataset consists of the best $2$ trajectories from the offline dataset and $2$ random trajectories, and trajectory's length is $100$. The offline dataset also serves as the target dataset.

\subsection{Implementation Details}

\ourmethod does not rely on the goal information, so we set \texttt{\_partially\_observable=True} to remove the goal information in the state vector for Meta-World~\citep{yu2020meta} tasks. For all task pairs, we first perform K-means clustering and categorize trajectories segments in the feedback buffer into $2$ categories, setting $N=4$. Then we sample $2$ trajectories from each category and employ Algorithm~\ref{alg:CPA} to compute \ourmethod labels. CPA accuracy in Table~\ref{tab:main_results} is calculated based on the comparison between the ground-truth preference labels and the labels inferred by the CPA method. Specifically, for each pair of trajectories, we compare the preference label inferred by CPA with the corresponding ground-truth label. A prediction is correct if the CPA label matches the ground-truth label. The accuracy is then calculated as the ratio of correctly predicted labels to the total number of trajectory pair comparisons, formulated as:
\begin{equation*}
    \text{Accuracy}=\frac{N_{\text{correct}}}{N_{\text{total}}} \times 100\%,
\end{equation*}
where $N_{\text{correct}}$ denotes the number of trajectory pairs for which the CPA label accurately matches the ground-truth label, and $N_{\text{total}}$ represents the total number of trajectory pair comparisons made. The detailed hyperparameters of PT and IQL are presented in Table~\ref{tab:PT_hyperparameters} and Table~\ref{tab:IQL_hyperparameters}, respectively.

\begin{table*}[!htbp]
\centering
\caption{Hyperparameters of PT.}
\begin{tabular}{ll}
\toprule
\textbf{Hyperparameter} & \textbf{Value}  \\
\midrule
Number of layers & $1$ \\
Number of attention heads & $4$ \\
Embedding dimension & $256$ \\
Batch size & $256$ \\
Optimizer & AdamW \\
Optimizer betas & $(0.9, 0.99)$ \\
Learning rate & $0.0001$ \\
Learning rate decay & Cosine decay \\
Weight decay & $0.0001$ \\
Dropout & $0.1$ \\
\bottomrule
\end{tabular}
\label{tab:PT_hyperparameters}
\end{table*}

\begin{table*}[!htbp]
\centering
\caption{Hyperparameters of IQL.}
\begin{tabular}{ll}
\toprule
\textbf{Hyperparameter} & \textbf{Value}  \\
\midrule
Network (Actor, Critic, Value network) & $(256, 256)$ \\
Optimizer (Actor, Critic, Value network) & Adam \\
Learning rate (Actor, Critic, Value network) & $0.0003$ \\
Discount & $0.99$ \\
Temperature & $3.0$ (Meta-World), $0.5$ (Robomimic) \\
Expectile & $0.7$ \\
Dropout & None (Meta-World), $0.1$ (Robomimic) \\
Soft target update rate & $0.005$ \\
\bottomrule
\end{tabular}
\label{tab:IQL_hyperparameters}
\end{table*}

\section{Additional Results}
\label{app:results}

In this section, we conduct additional experiments to evaluate the sensitivity of our method to several critical hyperparameters, which include the robust term's weight coefficient $\lambda$ in $\gL_{\text{ce}}$, the regularization weight coefficient $\alpha$, and the RPT margin $\eta$. The following experiments use Button Press as the source task for Sweep Into, Faucet Close for Window Open, and Square-MH for Lift-MH.

\paragraph{Robust Term's Weight Coefficient $\lambda$ in $\gL_{\text{ce}}$.}

\begin{figure*}[!ht]
\centering
\begin{tabular}{ccc}
\subfloat[Sweep Into]{\includegraphics[width=0.28\linewidth]{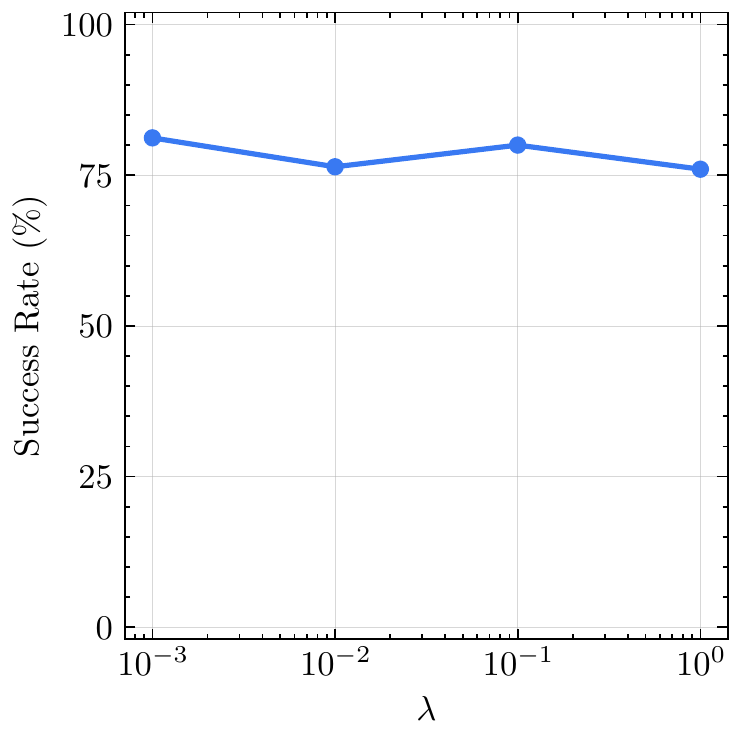}}
& \subfloat[Window Open]{\includegraphics[width=0.28\linewidth]{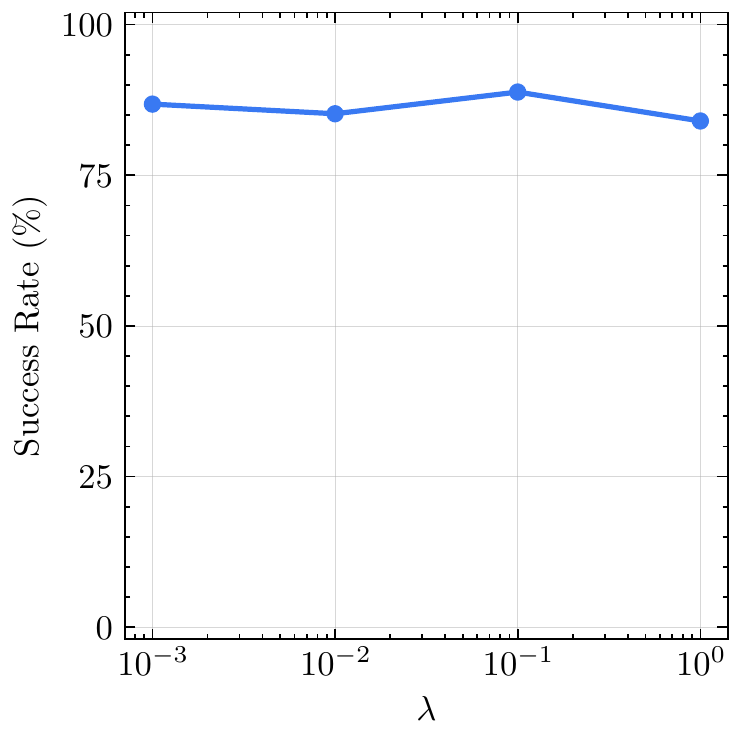}}
& \subfloat[Lift-MH]{\includegraphics[width=0.28\linewidth]{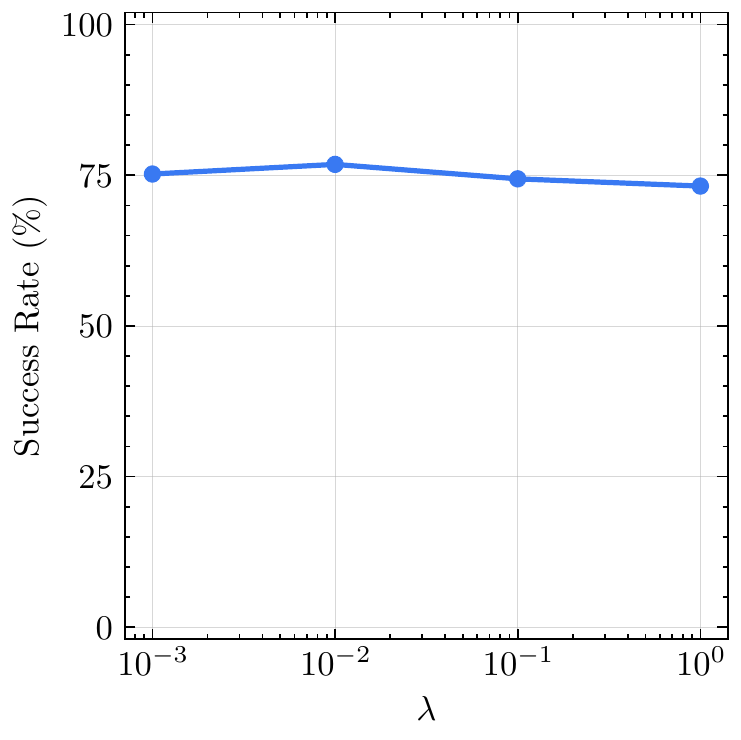}}
\end{tabular}
\caption{Success rate of Sweep Into, Window Open and Lift-MH tasks across different $\lambda$ values.}
\label{fig:abla_lambda}
\end{figure*}

The hyperparameter $\lambda$ balances the effects of mean and sampled rewards in~\Eqref{eq:robust_sup_loss}. To assess the sensitivity of our method to the weight $\lambda$, we perform supplementary experiments with different $\lambda = \{0.001, 0.01, 0.1, 1\}$. The results in Figure~\ref{fig:abla_lambda} demonstrate our method's robustness against $\lambda$ variations.

\paragraph{Regularization Weight Coefficient $\alpha$.}

\begin{figure*}[!ht]
\centering
\begin{tabular}{ccc}
\subfloat[Sweep Into]{\includegraphics[width=0.28\linewidth]{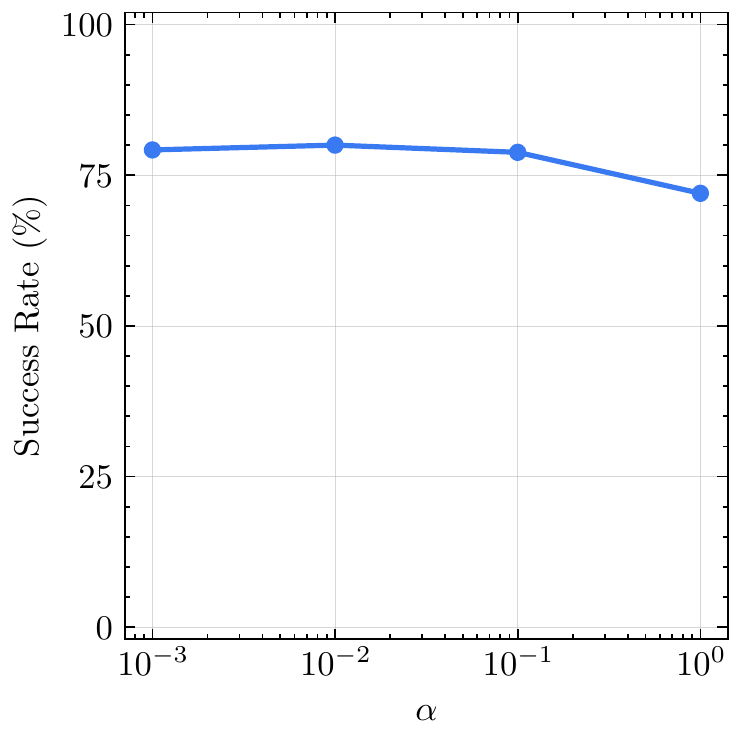}}
& \subfloat[Window Open]{\includegraphics[width=0.28\linewidth]{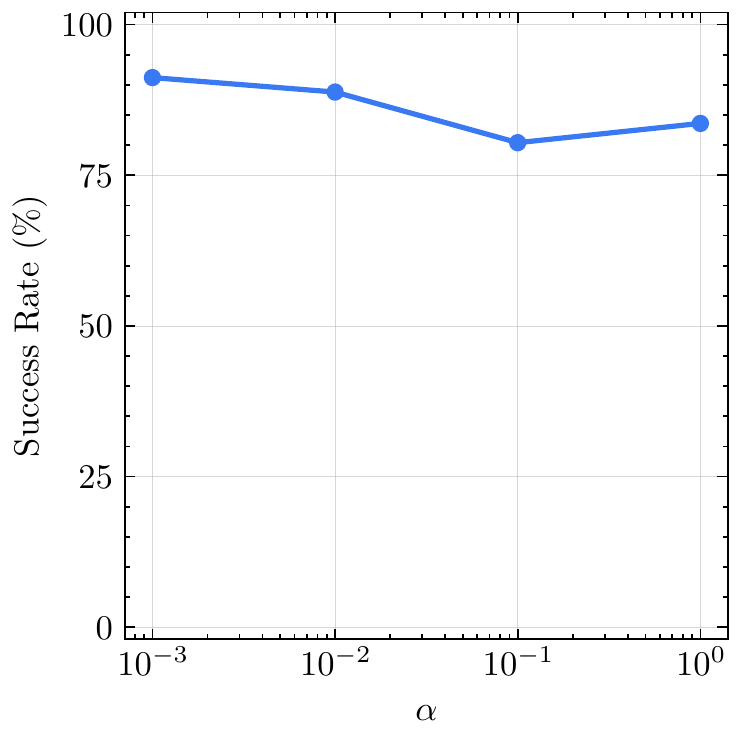}}
& \subfloat[Lift-MH]{\includegraphics[width=0.28\linewidth]{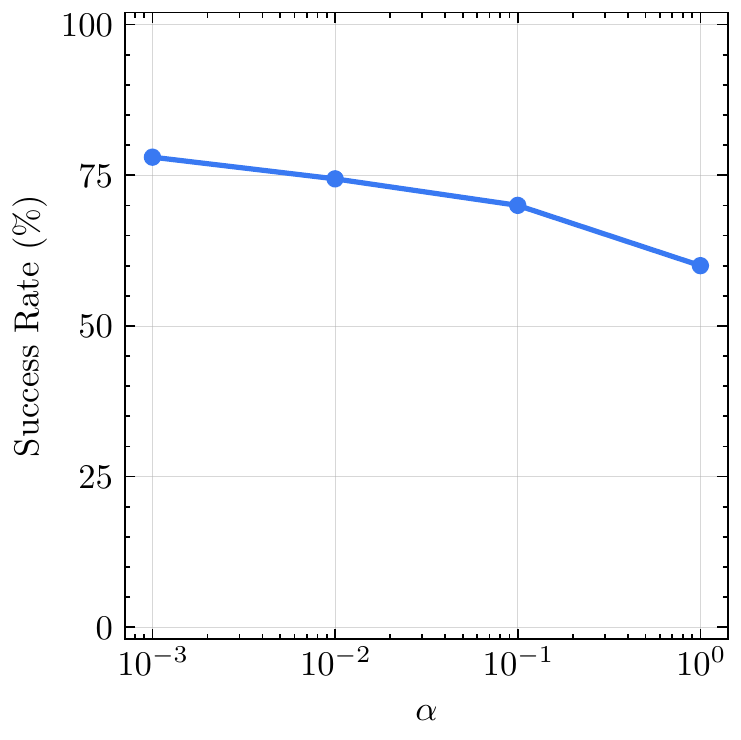}}
\end{tabular}
\caption{Success rate of Sweep Into, Window Open, and Lift-MH tasks across different $\alpha$ values.}
\label{fig:abla_alpha}
\end{figure*}

To examine the influence of the weight coefficient $\alpha$ in~\Eqref{eq:total_loss}, we evaluate our approach with $\alpha = \{0.001, 0.01, 0.1, 1\}$. As Figure~\ref{fig:abla_alpha} shows, our method retains high success rate with small $\alpha$ values. Conversely, a larger $\alpha$ slightly reduce the performance, as it diminish the contribution of $\gL_{\text{ce}}$ to reward learning, which further affects the accuracy of the reward function.

\paragraph{RPT Margin $\eta$.}

\begin{figure*}[!ht]
\centering
\begin{tabular}{ccc}
\subfloat[Sweep Into]{\includegraphics[width=0.28\linewidth]{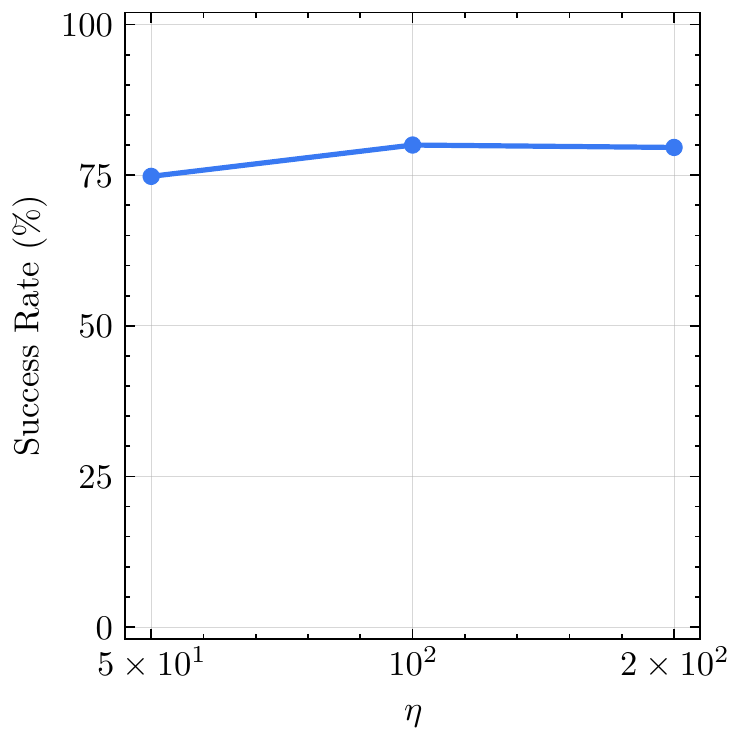}}
& \subfloat[Window Open]{\includegraphics[width=0.28\linewidth]{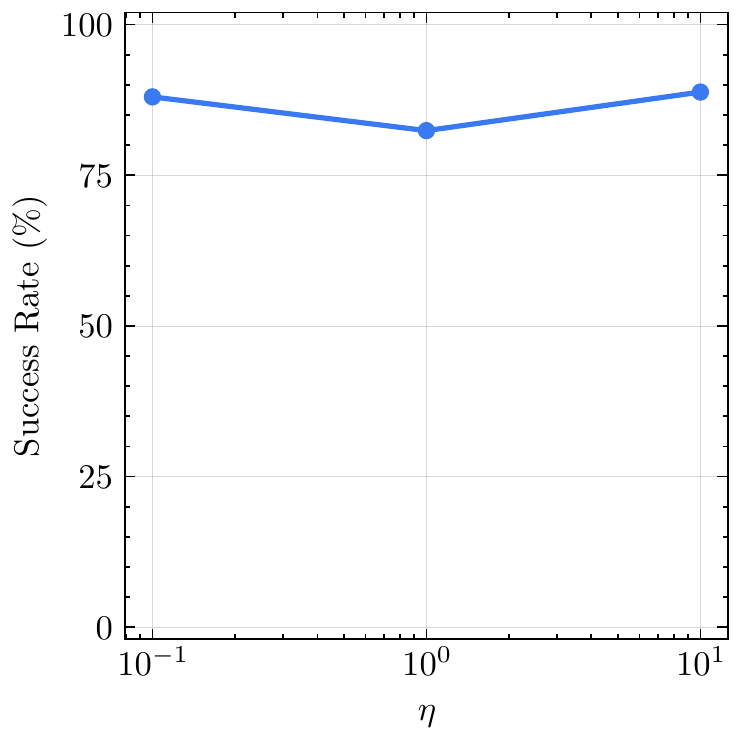}}
& \subfloat[Lift-MH]{\includegraphics[width=0.28\linewidth]{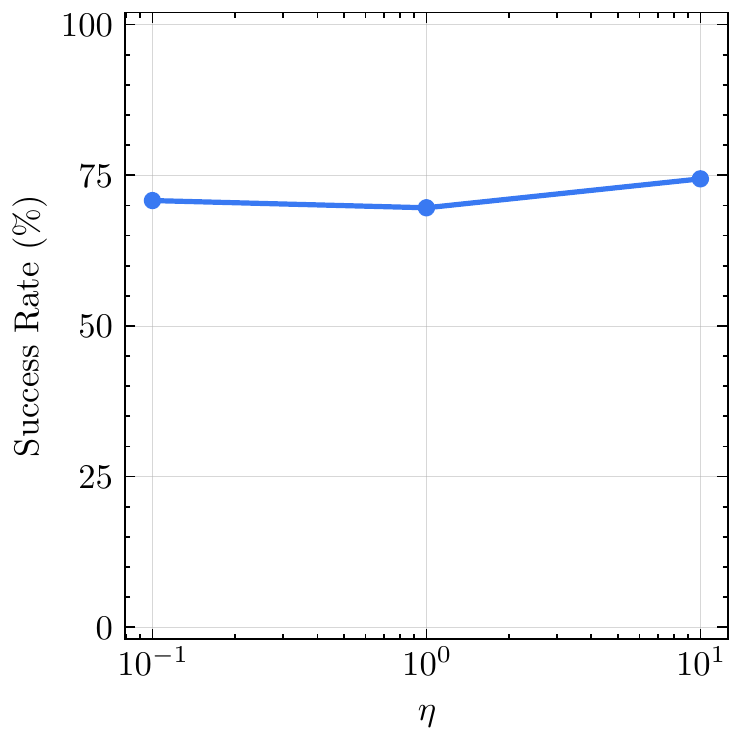}}
\end{tabular}
\caption{Success rate of Sweep Into, Window Open, and Lift-MH tasks across different $\eta$ values.}
\label{fig:abla_eta}
\end{figure*}

$\eta$ serves as a variance constraint in~\Eqref{eq:reg_loss}. Further experiments are conducted to evaluate this parameter's influence. For Sweep Into, $\eta = \{50, 100, 200\}$ are used for evaluation, while $\eta = \{0.1, 1, 10\}$ are used for Window Open and Lift-MH tasks. The results in Figure~\ref{fig:abla_eta} demonstrate that our method is not sensitive to the changes of $\eta$.

\paragraph{Number of Source Task Trajectories $M$.}

We additionally conduct experiments to evaluate the effect of the number of source task trajectories to our method, across $M \in \{4, 8, 16\}$. The results in Table~\ref{tab:abla_src_num} show that our method is not sensitive to the number of source trajectories.

\begin{table*}[!ht]
\centering
\small
\caption{Success rate and accuracy of \CPA labels across different numbers of source task trajectories. The results are reported with mean and standard deviation of success rate across five runs.}
\resizebox{0.95\linewidth}{!}{
\begin{tabular}{llrrrrrr}
    \toprule
    \multirow{2}{*}{\bf Source Task} & \multirow{2}{*}{\bf Target Task} &  \multicolumn{2}{c}{\bf $M=4$} & \multicolumn{2}{c}{\bf $M=8$} & \multicolumn{2}{c}{\bf $M=16$} \\
    \cmidrule(l){3-4}\cmidrule(l){5-6}\cmidrule(l){7-8}
    & & \multicolumn{1}{c}{\bf RPT+\CPA} & \multicolumn{1}{c}{\bf \CPA Acc.} & \multicolumn{1}{c}{\bf RPT+\CPA} & \multicolumn{1}{c}{\bf \CPA Acc.} & \multicolumn{1}{c}{\bf RPT+\CPA} & \multicolumn{1}{c}{\bf \CPA Acc.} \\
    \midrule
    Button Press & Drawer Open  
    & \mean{84.0} \std{16.0}
    & \mean{76.6}
    & \mean{82.6} \std{17.7}
    & \mean{77.6}
    & \mean{85.2} \std{15.8}
    & \mean{76.6} \\
    Faucet Close & Window Open 
    & \mean{88.8} \std{6.7}
    & \mean{87.0}
    & \mean{85.2} \std{9.4}
    & \mean{85.0}
    & \mean{88.4} \std{11.0}
    & \mean{87.0} \\
    \midrule
    \multicolumn{2}{c}{\bf Average} & \mean{86.4} & \mean{81.8} & \mean{83.9} & \mean{81.3} & \mean{86.8} & \mean{81.8} \\
    \bottomrule
\end{tabular}
}
\label{tab:abla_src_num}
\end{table*}

\section{Discussion}

\subsection{How does the task similarity influence CPA?}
\label{app:discuss_when}

We employ two distinct metrics to concretely evaluate task similarity: reconstruction error~\citep{ammar2014automated} and latent distance. The former employs the Jensen-Shannon distance (JSD) to measure task similarity. The latter involves training a Variational Autoencoder (VAE)~\citep{VAE} on each task and measuring the Euclidean distance between the latent vectors of different tasks. The results summarized in Table~\ref{tab:app_recon} reveal that tasks with greater similarity (i.e., lower reconstruction error) exhibit higher cross-task preference transfer accuracy. The Pearson correlation coefficient between CPA Accuracy and Reconstruction Error is $-0.43$, indicating a negative linear relationship. This suggests that as the Reconstruction Error decreases, the CPA Accuracy tends to increase. Similarly, the Pearson correlation coefficient between CPA Accuracy and Latent Distance is $-0.67$, indicating a stronger negative linear relationship compared to the Reconstruction Error. This means that as the Latent Distance decreases, indicating greater task similarity, the CPA Accuracy tends to increase more strongly. These correlations support the idea that greater task similarity (as measured by lower Reconstruction Error and lower Latent Distance) is associated with higher CPA Accuracy. Such findings suggest that task similarity, as quantified by reconstruction error in this context, is a critical factor in the successful application of \ourmethod for preference transfer. Notably, while \ourmethod shows promising results in scenarios where the source and target tasks share considerable similarities, its performance diminishes in the face of substantial domain gaps, such as tasks in D4RL Gym Locomotion~\citep{D4RL}.

\begin{table*}[!htbp]
\centering
\caption{Reconstruction error and latent distance between evaluated task pairs.}
\resizebox{0.75\linewidth}{!}{
\begin{tabular}{ll|rrr}
    \toprule
    {\bf Source Task} & {\bf Target Task}
    & {\bf Reconstruction Error} & {\bf Latent Distance}
    & {\bf \CPA Acc.} \\
    \midrule
    \multirow{4}{*}{Button Press} & Window Open
    & \mean{5437.5} & \mean{2.19} & \mean{87.0} \\
     & Door Close
    & \mean{5446.1} & \mean{2.66} & \mean{78.0} \\
     & Drawer Open
    & \mean{5262.3} & \mean{2.54} & \mean{76.6} \\
     & Sweep Into
    & \mean{5656.1} & \mean{2.52} & \mean{69.5} \\
    \midrule
    \multirow{4}{*}{Faucet Close} & Window Open
    & \mean{5491.3} & \mean{2.23} & \mean{87.0} \\
     & Door Close
    & \mean{5443.3} & \mean{2.70} & \mean{72.0} \\
     & Drawer Open
    & \mean{5292.0} & \mean{2.54} & \mean{77.0} \\
     & Sweep Into
    & \mean{5691.6} & \mean{2.55} & \mean{68.4} \\
    \midrule
    \multirow{2}{*}{Square-MH} & Can-MH    
    & \mean{-} & \mean{3.82} & \mean{70.0} \\
     & Lift-MH   
    & \mean{-} & \mean{3.37} & \mean{63.2} \\
    \bottomrule
\end{tabular}
}
\label{tab:app_recon}
\end{table*}

\subsection{Does CPA work without a full ranking of all trajectory pairs?}

CPA does not inherently require a full ranking of all trajectory pairs from the source task. The sparsity of the alignment matrix $\matA^{jj^\prime}$ is indeed a feature that allows our method to operate effectively even when some pairwise preferences are unavailable. For instance, in the scenario shown in Figure 1 with $j=1, j^\prime=2$, where $\matT_{\cdot 1}=(0, 0, 0.25, 0)^\top$ and $\matT_{\cdot 2}^\top=(0, 0, 0, 0.25)^\top$. the resulting $\matA^{12}$ matrix demonstrates significant sparsity:
\begin{equation*}
    \matA^{12} = \left( \begin{array}{cccc} 0 & 0 & 0 & 0 \\ 0 & 0 & 0 & 0 \\ 0 & 0 & 0 & 0.0625 \\ 0 & 0 & 0 & 0 \\ \end{array} \right).
\end{equation*}
This example underscores that our approach can indeed function with sparse and incomplete pairwise preference information. The presence of non-zero elements in $\matA^{jj^\prime}$ reflects the method's capability to focus on the most informative trajectory pairings, thereby mitigating the need for exhaustive preference annotations. In cases where the most similar trajectory pairs lack explicit preference labels, our method can adapt by checking whether the elements of $\matA^{jj^\prime}$ are all $0$.

\subsection{Standard Deviation of \ourmethod.}

As shown in Table~\ref{tab:main_results}, the standard deviation of \ourmethod is higher than that of PT and PT+Semi, the reason is that \ourmethod achieves excellent results with some random seeds but performs less optimally with other seeds. This variability can indeed be attributed to the inherent stochasticity of reinforcement learning environments and the zero-shot nature of our approach.

\end{document}